\documentclass{article}

\usepackage[square,numbers]{natbib}
\PassOptionsToPackage{options}{natbib}

\usepackage[final]{neurips_2024}

\usepackage[utf8]{inputenc} % allow utf-8 input
\usepackage[T1]{fontenc}    % use 8-bit T1 fonts
\usepackage{hyperref}       % hyperlinks
\usepackage{url}            % simple URL typesetting
\usepackage{booktabs}       % professional-quality tables
\usepackage{amsfonts}       % blackboard math symbols
\usepackage{nicefrac}       % compact symbols for 1/2, etc.
\usepackage{microtype}      % microtypography
\usepackage{xcolor}         % colors
\usepackage{amsmath}

\usepackage{graphicx}
\usepackage{wrapfig}
\usepackage{subcaption}

\usepackage{enumitem}

\usepackage{tabularx,colortbl}

\usepackage{multirow}

\usepackage[toc,page,header]{appendix}

\definecolor{calpolypomonagreen}{rgb}{0.12, 0.3, 0.17}
\hypersetup{
    colorlinks=true,
    citecolor=calpolypomonagreen,
    linkcolor=calpolypomonagreen,
    filecolor=magenta,      
    urlcolor=magenta,
    pdftitle={Overleaf Example},
    pdfpagemode=FullScreen,
}
	
\definecolor{ABC}{rgb}{0.9,0.9,0.9}

\newcolumntype{?}{!{\vrule width 1.5pt}}

\title{Discrete Dictionary-based Decomposition Layer \\ for Structured Representation Learning}

\author{%
  Taewon Park$^1$ ~~~~ Hyun-Chul Kim$^1$ ~~~~ Minho Lee$^{1,2}$ \\
  $^1$Kyungpook National University, South Korea \\
  $^2$ALI Co., Ltd., South Korea \\
  \texttt{ptw7998@gmail.com},~ \texttt{hyunchul\_kim@knu.ac.kr},~ \texttt{mholee@gmail.com} \\
}

\begin{document}

\maketitle

\begin{abstract}
Neuro-symbolic neural networks have been extensively studied to integrate symbolic operations with neural networks, thereby improving systematic generalization. Specifically, Tensor Product Representation (TPR) framework enables neural networks to perform differentiable symbolic operations by encoding the symbolic structure of data within vector spaces. However, TPR-based neural networks often struggle to decompose unseen data into structured TPR representations, undermining their symbolic operations. To address this decomposition problem, we propose a \textbf{D}iscrete \textbf{D}ictionary-based \textbf{D}ecomposition (\texttt{D3}) layer designed to enhance the decomposition capabilities of TPR-based models. \texttt{D3} employs discrete, learnable key-value dictionaries trained to capture symbolic features essential for decomposition operations. It leverages the prior knowledge acquired during training to generate structured TPR representations by mapping input data to pre-learned discrete features within these dictionaries. \texttt{D3} is a straightforward drop-in layer that can be seamlessly integrated into any TPR-based model without modifications. Our experimental results demonstrate that \texttt{D3} significantly improves the systematic generalization of various TPR-based models while requiring fewer additional parameters. Notably, \texttt{D3} outperforms baseline models on the synthetic task that demands the systematic decomposition of unseen combinatorial data.\footnote{The code of \texttt{D3} is publicly available at \href{https://github.com/taewonpark/D3}{https://github.com/taewonpark/D3}}
\end{abstract}

\section{Introduction}

Compositional generalization, aiming at understanding unseen data by combining known concepts, is essential for neural networks to handle complex tasks \cite{fodor1988connectionism,lake2017building,lake2018generalization,livska2018memorize,hupkes2020compositionality,greff2020binding}. Tensor Product Representation (TPR) framework \cite{smolensky1990tensor} facilitates this by embedding the symbolic structure of data within vector spaces, providing neural networks with compositional capabilities. Within this framework, individual objects are decomposed at the representation level into distinct symbolic components called \textit{role-filler} pairs\footnote{The \textit{roles} and \textit{fillers} depend on the task at hand. For example, in a tree structure, the \textit{role} corresponds to a position within the tree, while the \textit{filler} represents the label associated with that position \cite{soulos2023differentiable}. In associative memory, the \textit{role} is analogous to an associative key, and the \textit{filler} corresponds to the associative value \cite{schlag2018learning,schlag2020learning}.}. The TPR framework encodes each object by taking a tensor product of its \textit{role} vector and \textit{filler} vector, represented as $T = filler \otimes role$, and then superimposes them to represent multiple objects within a single representation. During decoding, the TPR framework retrieves specific \textit{fillers}—essential for solving tasks—from the superimposed representation through matrix multiplication using an \textit{unbinding operator} correlated to a particular \textit{role}, $filler=T \cdot unbind$. This retrieved \textit{filler} is then utilized in downstream tasks. Based on this property, TPR-based neural networks have demonstrated significant generalization and applicability in fields such as associative reasoning \cite{schlag2018learning,schlag2020learning}, mathematical problem-solving \cite{schlag2019enhancing}, and natural language processing \cite{jiang2021enriching,shi2022stepgame,palangi2018question,soulos2023differentiable}.

Despite their successes, the TPR-based approaches pose a significant challenge known as a decomposition problem \cite{smolensky1990tensor,park2023attention}, which refers to the difficulty of decomposing input data into TPR components, such as \textit{roles}, \textit{fillers}, and \textit{unbinding operators}. Without accurate decomposition, TPR-based models fail to represent the symbolic structure of data, causing a decline in the performance of the TPR operations. Recently, inspired by an object-centric learning method \cite{locatello2020object}, \citet{park2023attention} proposes an attention-based iterative decomposition (AID) module to address this issue. AID uses competitive attention to iteratively refine structured representations, thereby enhancing the systematic generalization of TPR-based models. However, it still struggles to generalize all possible combinations of known symbols in simple synthetic tasks. This failure is likely attributable to its insufficient mechanism for explicitly mapping input data to known symbolic features observed during training. Therefore, the decomposition module may need an additional mechanism to store observed symbolic features during training and utilize it to effectively decompose unseen combinatorial data of known symbols.

In another line of work, discrete representation learning has been explored to improve the efficiency, interpretability, and generalization capabilities of neural networks \cite{van2017neural,lample2019large,liu2021discrete,tamkin2023codebook,hsu2024disentanglement}. This approach involves mapping continuous input data into discrete representations by finding the nearest features in a predefined codebook. The features within the codebook are learnable parameters, specifically trained to capture the latent features of data during training phase \cite{van2017neural}. Some researchers have applied discrete representation techniques to extract specific types of representations from unstructured data \cite{kori2023grounded,wu2024structured,zhuang2024learning}. Other researchers have integrated discrete symbolic embeddings within the TPR framework to improve its interpretability \cite{palangi2018question,jiang2021enriching}. However, these methods are designed for specific applications, such as question-answering and summarization tasks, making them difficult to integrate into other TPR-based models.

\begin{figure}[t]
\begin{center}
%\vskip -0.3in
\includegraphics[width=0.95\columnwidth]{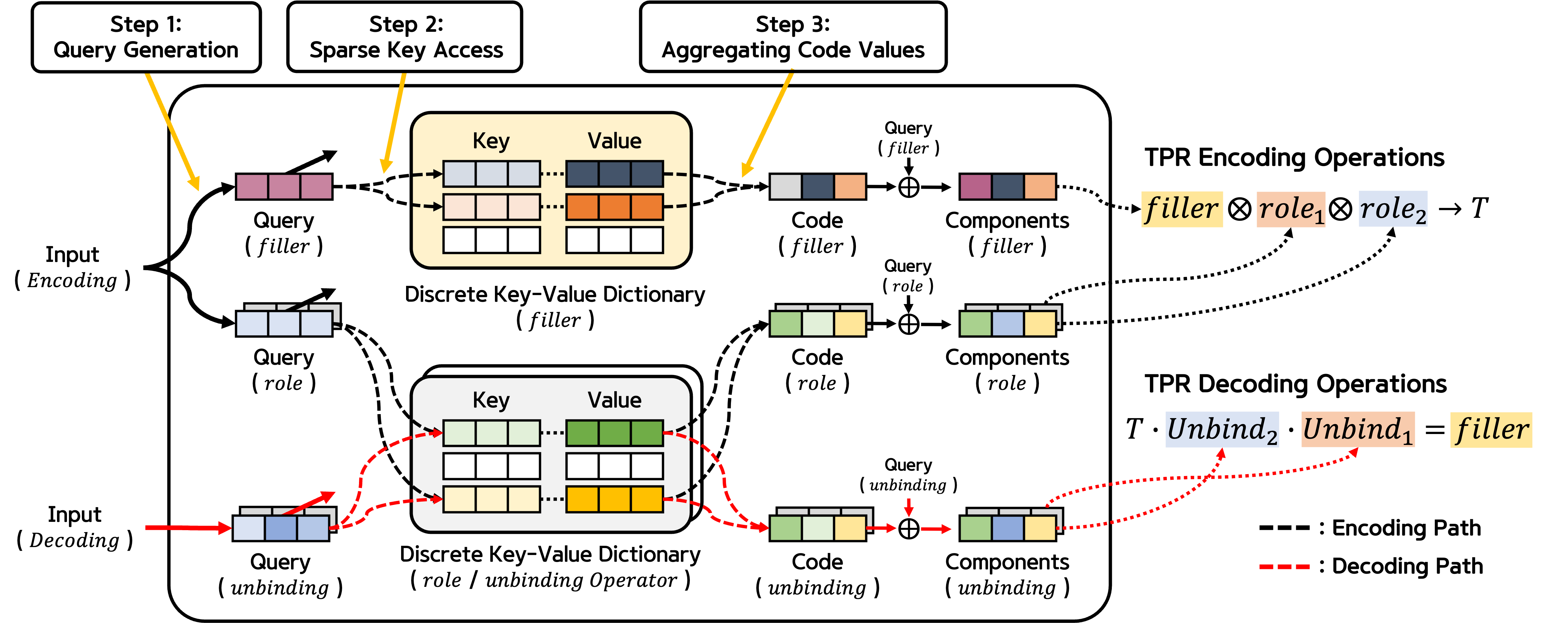}
%\vskip -0.05in
\vskip 0.05in
\caption{\textbf{Overview of D3.}
\texttt{D3} generates structured TPR representations by mapping input data to the nearest pre-learned symbolic features stored within discrete, learnable dictionaries. Each dictionary is linked explicitly to specific TPR components, such as \textit{roles}, \textit{filler}, and \textit{unbinding operators}. Notably, \texttt{D3} uses a shared dictionary configuration between the \textit{roles} and \textit{unbinding operators}. This figure illustrates, for example, that $\textit{role}_1$ and $\textit{unbind}_1$ share one dictionary, while $\textit{role}_2$ and $\textit{unbind}_2$ share another. $T$ denotes a superimposed representation that represents multiple objects.
}
\vskip -0.25in
\label{fig.model}
\end{center}
\end{figure}

In this work, we propose a \textbf{D}iscrete \textbf{D}ictionary-based \textbf{D}ecomposition (\texttt{D3}) layer for structured representation learning within the TPR framework. \texttt{D3} employs the discrete representations techniques to utilize prior knowledge acquired during training for decomposition operations. Inspired by prior discrete key-value architectures \cite{lample2019large,trauble2023discrete}, \texttt{D3} consists of multiple dictionaries, each comprising discrete, learnable key-value pairs. Unlike prior work, each dictionary of \texttt{D3} is linked explicitly to individual TPR components, such as \textit{role}, \textit{filler}, and \textit{unbinding operator}. This design allows each dictionary to capture and store the discrete features of its corresponding TPR components during training. \texttt{D3} acts as a drop-in layer that maps input data into pre-learned discrete features for the decomposition of TPR components through a three-step process, as illustrated in Fig.~\ref{fig.model}. First, it generates multiple queries from the input data, with each query utilized for different TPR components. Next, it identifies the nearest codebook keys within each dictionary based on these queries. Finally, \texttt{D3} generates structured TPR representations by aggregating the codebook values corresponding to these keys. Moreover, \texttt{D3} can be seamlessly integrated into any TPR-based model by replacing the TPR component generation layer without requiring further modifications.

\textbf{Our main contributions} are as follows.
\vspace{-\topsep}
\begin{itemize}[leftmargin=0.5cm]
    \item We propose a novel \texttt{D3} layer to tackle the decomposition problem inherent in the TPR-based approaches. \texttt{D3} leverages discrete, learnable dictionaries to enhance the decomposition capabilities of TPR-based models. By mapping input data to pre-learned discrete features stored within the dictionaries, \texttt{D3} effectively generates structured TPR representations.
    \item We conduct extensive experiments across various systematic generalization tasks, including synthetic associative recall and text/visual question-answering tasks. Our experimental results show that \texttt{D3} significantly enhances the generalization performance of TPR-based models, demonstrating its effectiveness on systematic generalization tasks.
    \item Our analyses show that \texttt{D3} generates well-bound structured representations that are satisfactory for the requirements of the TPR framework, utilizing the discrete, learnable dictionaries.
\end{itemize}

\section{Related Work}
\label{sec.related}

\paragraph{Decomposition Problem.}
Compositional generalization in neural networks, which allows for generalizing beyond training data, has been extensively studied \cite{fodor1988connectionism,lake2017building,lake2018generalization,livska2018memorize,hupkes2020compositionality,greff2020binding,webb2020emergent}. One important capability for achieving this is a \textit{segregation}, as discussed in \citet{greff2020binding}, which enables the formation of meaningful representations from structured and unstructured data \citep{goyal2019recurrent,locatello2020object}. TPR-based neural networks also rely on this capability to generate structured representations for TPR components such as \textit{roles}, \textit{fillers}, and \textit{unbinding operators}. In the TPR framework, these structured representations must satisfy specific conditions to ensure accurate encoding and decoding. First, \textit{roles} need to be linearly independent to avoid \textit{filler} overlap. Second, the \textit{unbinding operator} must correlate with the corresponding \textit{roles} to accurately retrieve associated \textit{fillers}. Recent work \cite{park2023attention} has shown that existing TPR-based models often fail to generate structured representations that meet these conditions, undermining their symbolic operations. To address this, an attention-based decomposition module \cite{park2023attention} has been introduced, but it still shows limited performance on synthetic tasks involving the decomposition of unseen combinatorial data. In this work, we address the decomposition problem within the TPR framework using a discrete dictionary-based method, advancing the research further.

\paragraph{Discrete Representation Learning.}
Discrete neural representation learning has introduced a codebook of discrete, learnable representations into neural networks \cite{van2017neural}. During training, each discrete representation captures underlying latent features by mapping continuous input data to the nearest features within the codebook, which are then used for downstream tasks. Recent work on object-centric learning has utilized discrete representations to extract specific types of features from unstructured data, leveraging latent features learned during training \cite{kori2023grounded,wu2024structured}. Some researchers have proposed a separate key-value codebook for learning discrete representations, demonstrating its effectiveness in systematic generalization \cite{liu2021discrete} and robustness against distributional shifts \cite{trauble2023discrete}. Inspired by these findings, we develop a separate key-value-based discrete dictionary method to enhance the decomposition capabilities of TPR-based models. Other researchers have introduced a discrete symbolic embedding layer to improve the interpretability of TPR-based models, showing the feasibility of discrete representations in the TPR framework \cite{palangi2018question,jiang2021enriching}. However, their methods focus on encoding processes and specific tasks such as question-answering \cite{palangi2018question} and abstractive summarization \cite{jiang2021enriching}. In contrast, our work addresses the decomposition problem in TPR-based approaches, and our \texttt{D3} method is a drop-in solution that can be easily adapted to any TPR-based model.

\paragraph{Memory Network.}
Research on memory networks has focused on enhancing neural network capacity by integrating external memory \cite{sukhbaatar2015end,graves2014neural,graves2016hybrid,rae2016scaling,santoro2018relational,webb2020emergent}. Memory-augmented neural networks store variable lengths of sequential data in this external memory and retrieve necessary information using various addressing methods \cite{sukhbaatar2015end,graves2016hybrid}. These writing and reading mechanisms share many similarities with our \texttt{D3} approach. However, while memory networks store input features sequentially in their memory states as a continuous stream, \texttt{D3} updates symbolic feature information through gradient descent into codebook parameters within dictionaries. This distinctive characteristic allows \texttt{D3} to leverage the learned discrete features to decompose unseen data after training. In another work, \citet{lample2019large} introduces a learnable key-value memory layer to improve the efficiency of the Transformer by replacing the feed-forward layer. Unlike their memory layer, \texttt{D3} employs key-value pairs in dictionaries explicitly linked to individual TPR components, making it well-suited for the TPR framework.

\section{Method}
\label{sec.method}

In this section, we explain how the \texttt{D3} module generates structured representations of the TPR components using discrete, learnable dictionaries. We then introduce configurations of \texttt{D3} and how it can be applied to our baseline models.

\subsection{Discrete Dictionary-based Decomposition module}

\texttt{D3} is a discrete dictionary-based drop-in layer designed to enhance the decomposition capabilities of TPR-based approaches. At every time step, \texttt{D3} decomposes input data into TPR components, such as \textit{roles}, \textit{fillers}, and \textit{unbinding operators}, by mapping input data to pre-learned symbolic features within dictionaries. These dictionaries consist of discrete, learnable codebook key-value pairs, denoted as $\{\mathcal{D}^j\}_{j=1}^{N\text{component}}$ as shown in Eq.~\ref{eq.dict}. Each dictionary $\mathcal{D}^j$ is explicitly linked to a $j$-th TPR component, allowing it to learn the symbolic features required for generating the specific TPR component. This design also enables the generation of structured representations for different TPR components individually and in parallel.
\begin{equation} \label{eq.dict}
\mathcal{D}^j:=\{(\textsf{k}_i^j,\textsf{v}_i^j)~|~\textsf{k}_i^j \in \mathbb{R}^{D_\text{query}},\textsf{v}_i^j \in \mathbb{R}^{D_\text{code}} \}_{i=1}^{N_\text{code}} ~~~~~ \text{where}~~ j=1,..., N_\text{component}    
\end{equation}
where $\mathcal{D}^j$ denotes the discrete, learnable dictionary for the $j$-th TPR component, $\textsf{k}$ denotes a learnable codebook key, and $\textsf{v}$ denotes a learnable codebook value. In the next paragraph, we describe how \texttt{D3} generates TPR components using these dictionaries in three steps.

\paragraph{Step 1: Query Generation.}

At each time step $t$, \texttt{D3} takes input data, denoted as $\texttt{input}_t \in \mathbb{R}^{D\text{input}}$, and generates the query, denoted as $\texttt{queries}_t \in \mathbb{R}^{N\text{component}  \times D_\text{input}}$, for each $j$-th TPR component using a query network, $f_\text{query}^j: \texttt{input}_t \mapsto \texttt{query}^j_t \in \mathbb{R}^{D{\text{query}}}$. The query network can be any neural network; in this study,  we use a feed-forward network with a single layer. Additionally, we apply a layer normalization \cite{ba2016layer} and a dropout of $p_\text{dropout}$ \cite{srivastava2014dropout} to $\texttt{query}^j_t$.

\paragraph{Step 2: Sparse Key Access.}

\texttt{D3} searches for the nearest keys from each dictionary, $\mathcal{D}^j$, based on the generated $\texttt{query}^j_t$. We measure the similarity using the inner product between $\texttt{query}^j_t$ and $\{ \textsf{k}_i^j \}_{i=1}^{N_\text{code}}$. Then, \texttt{D3} selects top-$k$ codebook keys in order of largest similarity, as follows.
\begin{equation}
\mathcal{I}^j = \mathcal{T}_k ( {\texttt{query}^j_t}^\top \hat{\textsf{k}}^j_i) ~~~~~ \text{where}~~ \hat{\textsf{k}}^j_i = \textsf{k}^j_i / ||\textsf{k}^j_i||_2   
\end{equation}
where $\mathcal{T}_k$ denotes the top-$k$ operator that finds the indices of $k$ largest values, and $\mathcal{I}^j$ denotes the indices of the $k$ most similar keys within $\mathcal{D}^j$. We found that applying $L2$ normalization to keys before the inner product mitigates the codebook collapse problem.

\paragraph{Step 3: Aggregation of Code Values.}

\texttt{D3} computes the normalized score for selected codebook keys, denoted as $w_t^j$, and aggregates codebook values corresponding to selected codebook keys with $w_t^j$, as follows.
\begin{equation}
\texttt{code}^j_t = \Sigma_{i \in \mathcal{I}} w^j_{t,i} \textsf{v}^j_i  ~~~~~ \text{where}~~ w^j_t = \text{Softmax}( {\texttt{query}^j_t}^\top \hat{\textsf{k}}^j_i))_{i \in \mathcal{I}^j}
\end{equation}
Then, \texttt{D3} maps $\texttt{query}^j_t$ to a dimension of $D_\text{code}$ and adds this projected vector to $\texttt{code}^j_t$. The summed vectors are mapped to a dimension of $D_\text{component}$ to generate structured representations of TPR components, as follows.
\begin{equation}
\texttt{component}_t^j = \texttt{code}^j_t + \text{layer}_\text{residual} (\texttt{query}^j_t) \in \mathbb{R}^{ D_\text{code}}
\end{equation}
\begin{equation}
\overline{\texttt{component}}_t^j = \text{layer}_\text{final} (\texttt{component}_t^j) \in \mathbb{R}^{ D_\text{component}}
\end{equation}

where $\text{layer}_\text{residual}$ and $\text{layer}_\text{final}$ denote a feed-forward network with a single layer. Those $\overline{\texttt{components}}_t$ are then utilized for TPR operations to solve the downstream tasks.

\subsection{Module Configurations}

In this section, we describe the configurations of \texttt{D3} when applied to TPR-based models.

\paragraph{Shared Dictionary between Role and Unbinding Operator.}
As discussed in Section~\ref{sec.related}, \textit{roles} and \textit{unbinding operators} should have correlated features for accurate TPR operations. Considering this characteristic of the TPR framework, we share the dictionaries of \textit{roles} and \textit{unbinding operators}. This shared dictionary also reduces the number of learnable parameters.

\paragraph{D3 Applied to Filler.}
While the TPR framework requires specific conditions for \textit{roles} and \textit{unbinding operators} for accurate TPR operations, there are no such requirements for \textit{fillers}. Therefore, we explore two configurations in this study: applying \texttt{D3} to generate \textit{fillers} (\textit{w/ F}) and not applying \texttt{D3} to generate fillers (\textit{w/o F}). In the \textit{w/o F} configuration, we follow the baseline models to generate the \textit{filler} representations.

\subsection{Integration of \texttt{D3} into Existing TPR-based Models}

In this section, we introduce our baseline models and explain how \texttt{D3} is applied to them, considering the configurations of \texttt{D3}. We use three TPR-based models as our baselines: FWM \cite{schlag2020learning}, TPR-RNN \cite{schlag2018learning}, and Linear Transformer \cite{katharopoulos2020transformers}. Notably, integrating \texttt{D3} into these baseline models requires only substituting their TPR component generation layer with \texttt{D3} without further modifications.

\paragraph{Fast Weight Memory.}

Fast Weight Memory (FWM) \cite{schlag2020learning} is a TPR-based memory network designed for understanding long sequential contexts. It proposes a single word-level TPR operation related to the perceptron learning rule \cite{rosenblatt1958perceptron}. It has shown significant associative reasoning capability in reinforcement learning and natural language processing tasks. FWM requires two types of \textit{roles} ($\textit{role}_1$ and $\textit{role}_2$) and one \textit{filler} for encoding, as well as two types of \textit{unbinding operators} ($\textit{unbind}_1$ and $\textit{unbind}_2$) for decoding. When \texttt{D3} is integrated into FWM, it employs three dictionaries for the shared dictionary configuration: one for the $\textit{role}_1$ and $\textit{unbind}_1$, another for the $\textit{role}_2$ and $\textit{unbind}_2$, and the other for \textit{filler}, as shown in Fig.~\ref{fig.model}.

\paragraph{TPR-RNN.}

TPR-RNN \cite{schlag2018learning} is a sentence-level TPR-based memory network designed for basic text question-answering tasks \cite{weston2015towards}. It incorporates various encoding operations such as writing, moving, and backlink to process sequential data at the sentence level. These operations necessitate different encoding components with varying dimensions, making direct connections to the decoding components challenging. As a result, we do not apply the shared dictionary configuration to TPR-RNN; instead, we use a shared query network without layer normalization. Furthermore, due to the differing dimensions of the TPR components in TPR-RNN, we employ distinct $\text{layer}_\text{final}$ layers for each TPR component.

\paragraph{Linear Transformer.}

Linear Transformer \cite{katharopoulos2020transformers} linearizes the attention mechanism to improve the computational efficiency of the Transformer \cite{vaswani2017attention}. Recently, \citet{schlag2021linear} demonstrated the equivalence between TPR and the linear attention mechanism, indicating that the key, value, and query in linear attention correspond to the \textit{role}, \textit{filler}, and \textit{unbinding operator}, respectively. Building on this work, we apply \texttt{D3} to generate the query, key, and value in the Linear Transformer. Unlike TPR-RNN and FWM, the Linear Transformer utilizes multi-head operations. Therefore, we use distinct dictionaries for each head, with the key and query of each head sharing the same dictionary.

\section{Experiment}

In this section, we evaluate the effectiveness of \texttt{D3} across various tasks, including a synthetic task, text/visual question-answering tasks, and a language modeling task. To assess the decomposition capabilities, we follow the experimental settings of the AID \cite{park2023attention}, a prior work addressing the decomposition problem in the TPR framework, and closely compare our \texttt{D3} model to baseline models and AID.

\subsection{Task}

\paragraph{Systematic Associative Recall (SAR) task.}
This task evaluates systematic generalization in memorizing and recalling combinatorial data \cite{park2023attention}. It consists of a discovery phase and an inference phase. During the discovery phase, the model receives the combinatorial sequential items, each combining two symbols, $x \in X$ and $y \in Y$ where $X=X_1 \cup X_2 \cup X_3$ and $Y=Y_1 \cup Y_2$. The model is then required to predict an associated $y$  when a specific $x$ is presented. The SAR task uses different combination settings between training and evaluation to target systematic generalization specifically. During training, the model learns the following combination settings: (1) $X_1$ and $Y_1$, (2) $X_2$ and $Y_2$, and (3) $X_3$ and $Y$. At the evaluation, on the other hand, the model should generalize unseen combination settings, specifically $X_1$ and $Y_2$. Additionally, the task includes a hyper-parameter $p=\frac{|X_3|}{|X_2|+|X_3|}$ where $|X_i|$ denotes the cardinality of set $X_i$. By adjusting $p$, this task tests the systematic generalization of models under varying levels of exposure to different symbol combinations during training. In our study, we focus solely on the most challenging setting of the SAR task ($p=0.0$), where the subset $X_3$ is excluded. In the SAR task, the TPR framework regards $x$ as the \textit{role} and the \textit{unbinding operator}, and $y$ as the \textit{filler}. Therefore, TPR-based models should systematically decompose the combinatorial data into structured representations by mapping $x$ to the \textit{role} and $y$ to the \textit{filler} during the discovery phase, and mapping $x$ to the \textit{unbinding operator} during the inference phase to solve this task.

\paragraph{Systematic bAbI (sys-bAbI) task.}
This task is a variant of the bAbI task \cite{weston2015towards} designed to evaluate systematic generalization in text understanding and reasoning \cite{park2023attention}. It consists of 20 distinct sub-tasks, each comprising stories, relevant queries, and corresponding answers. The sys-bAbI task requires the models to remember the stories and predict corresponding answers to the queries. Unlike the original bAbI task, the sys-bAbI task evaluates the models with two aspects: (a) in-distribution (\textit{w/o sys diff}) and (b) with the systematic difference (\textit{w/ sys diff}) where each sub-task includes unseen words during training. Therefore, the models should learn task-independent text understanding to solve the sys-bAbI task.

\paragraph{Sort-of-CLEVR task.}
This task \cite{santoro2017simple} evaluates compositional generalization in visual relational reasoning. It consists of scene images, queries, and corresponding answers. This task requires the models to understand the properties of individual objects (\textit{Unary}) or the relationships between multiple objects (\textit{Binary} or \textit{Ternary}) within visual scene images, and predict the correct answers to the queries \cite{mittal2021compositional}. Therefore, the model should capture relationships within each object and between objects to solve this task.

\paragraph{WikiText-103 task.}
This task \cite{merity2016pointer} is a language modeling dataset consisting of lengthy corpora from Wikipedia. Although the WikiText-103 task does not directly measure the systematic generalization of the models, it is used to evaluate the effectiveness and applicability of \texttt{D3} on a large-scale task beyond relatively simple tasks.

\subsection{Experimental Results}
In this section, we present the experimental results of the SAR task, sys-bAbI task, sort-of-CLEVR task, and WikiText-103 task. In our experiments, we set  $D_\text{query}$ as $D_\text{code}/2$.

\subsubsection{TPR-based Memory Networks}

\begin{wrapfigure}{r}{0.5\textwidth}
\vskip -0.2in
    \centering
    \includegraphics[width=\linewidth]{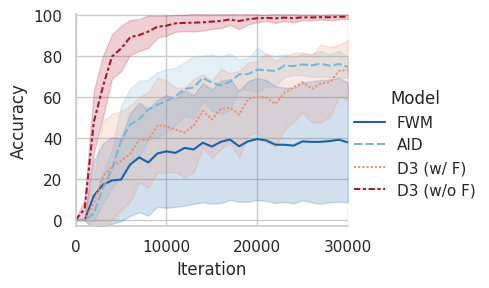}
    \label{fig.sar}
\vskip -0.2in
\caption{Test accuracy curve [\%] on the SAR task for 10 seeds, with shadowed area indicating SD.}
\label{fig.sar}
\end{wrapfigure}

First, we evaluate FWM with \texttt{D3} on the SAR task, which requires understanding the composition of two types of symbols, $x$ and $y$. TPR-based models are expected to solve this task perfectly by mapping each symbol to a specific TPR component during decomposition. However, as shown in Fig.~\ref{fig.sar}, FWM and AID fail to generalize unseen combinations of known symbols. In contrast, our \texttt{D3} module significantly outperforms other baseline models, achieving nearly 100\% accuracy. This result demonstrates that \texttt{D3} effectively decomposes unseen combinatorial data into TPR components using discrete dictionaries.

\begin{table}[!htb]
\centering
\begin{minipage}[t]{\textwidth}
\centering
    \caption{The mean word error rate [\%] on the sys-bAbI task for 10 seeds, with $\pm$ indicating SD.\label{table.babi}}

\begingroup
\setlength{\tabcolsep}{4pt} % Default value: 6pt
\renewcommand{\arraystretch}{1.2} % Default value: 1
\begin{center}
%{\scriptsize%\footnotesize
{\footnotesize
\begin{tabular}{l?l|l?c?c}
\hline

\hline

\hline
\bf Model            & \textit{w/o sys diff} ($\downarrow$)   &  \textit{w/ sys diff} ($\downarrow$)     & \textbf{Gap} ($\downarrow$)  & $\#$ params ($\downarrow$) \\ \hline

\hline

TPR-RNN & 0.79 \tiny{$\pm$ 0.16}   & 8.74 \tiny{$\pm$ 3.74}    & 7.95  & \textbf{0.14} $M$   \\

~~~~~~{+ \scriptsize{AID}} & \underline{0.69} \tiny{$\pm$ 0.08} & \underline{5.61} \tiny{$\pm$ 1.78}  & \underline{4.92}  & 0.32 $M$ \\
\rowcolor{ABC}
~~~~~~{+ \texttt{\textbf{D3}}} & \textbf{0.65} \tiny{$\pm$ 0.25}  & \textbf{3.50} \tiny{$\pm$ 2.07}  & \textbf{2.85} & \underline{0.17} $M$\\

\hline

\hline

FWM     & 0.79 \tiny{$\pm$ 0.14}   & 2.85 \tiny{$\pm$ 1.61}    & 2.06   & \textbf{0.73} $M$  \\

 ~~~~~~{+ \scriptsize{AID}} & \textbf{0.45} \tiny{$\pm$ 0.16}  & \textbf{1.21} \tiny{$\pm$ 0.66}  & \textbf{0.76}   & 1.23 $M$ \\
 
\rowcolor{ABC}
 ~~~~~~{+ \texttt{\textbf{D3}} (\textit{w/o F})} & 0.79 \tiny{$\pm$ 0.30}  & 2.58 \tiny{$\pm$ 1.12}  & 1.79  & \underline{0.75} $M$ \\

\rowcolor{ABC}
 ~~~~~~{+ \texttt{\textbf{D3}} (\textit{w/ F})} & \underline{0.75} \tiny{$\pm$ 0.17}  & \underline{1.96 }\tiny{$\pm$ 0.88}& \underline{1.21} & \underline{0.75} $M$ \\

 \hline

\hline
 
\hline
\end{tabular}}
\end{center}

\endgroup

\end{minipage}%

\medskip{}

\begin{minipage}[t]{\textwidth}

\centering
    \caption{The mean accuracy [\%] on the sort-of-CLEVR task for 10 seeds, with $\pm$ indicating SD.\label{table.clevr}}

\begingroup
\setlength{\tabcolsep}{4pt} % Default value: 6pt
\renewcommand{\arraystretch}{1.2} % Default value: 1

\begin{center}
{\footnotesize
\begin{tabular}{l?c?l|l|l?c}
\hline

\hline

\hline
 \bf Model       & $D_\text{code}$ & \textit{Unary} ($\uparrow$)~ & \textit{Binary} ($\uparrow$)  &  \textit{Ternary} ($\uparrow$)  & $\#$ params ($\downarrow$)   \\ \hline 
 
 \hline

Linear Transformer  &  - & 69.3 \tiny{$\pm$ 14.8}   & 75.5 \tiny{$\pm$ 1.3}   &  56.4 \tiny{$\pm$ 4.3} & \textbf{0.68} $M$  \\

~~~~~~~~~~{+ \scriptsize{AID}} &  - & \underline{98.9} \tiny{$\pm$ 0.2} & 78.6 \tiny{$\pm$ 0.3}  & \underline{63.7} \tiny{$\pm$ 1.2}  & 0.83 $M$ \\

\hline

\rowcolor{ABC}
~~~~~~~~~~{+ \texttt{\textbf{D3}} (\textit{w/o F})} & 128 & 73.9 \tiny{$\pm$ 16.5} & 77.2 \tiny{$\pm$ 2.2}  & 57.3 \tiny{$\pm$ 4.6}  & \underline{0.75} $M$  \\

\rowcolor{ABC}
 & 256 & 73.7 \tiny{$\pm$ 16.5}  & 77.8 \tiny{$\pm$ 2.5} & 57.9 \tiny{$\pm$ 5.8} & 0.96 $M$  \\

\hline

\rowcolor{ABC}
~~~~~~~~~~{+ \texttt{\textbf{D3}} (\textit{w/ F})} & 128 & \underline{98.9} \tiny{$\pm$ 0.2}  & \underline{79.5} \tiny{$\pm$ 0.8}  & 63.1 \tiny{$\pm$ 1.9}  & 0.80 $M$ \\

\rowcolor{ABC}
 & 256 & \textbf{99.0} \tiny{$\pm$ 0.3}  & \textbf{82.1} \tiny{$\pm$ 2.4}  & \textbf{68.8} \tiny{$\pm$ 1.2}  & 1.13 $M$  \\

 \hline
 
\hline

\hline
\end{tabular}}
\end{center}

\endgroup

\end{minipage}

\medskip{}

\begin{minipage}[t]{\textwidth}

\centering
    \caption{Perplexity on the WikiText-103 task.\label{table.wiki}}

\begingroup
\setlength{\tabcolsep}{4pt} % Default value: 6pt
\renewcommand{\arraystretch}{1.2} % Default value: 1

\begin{center}
{\footnotesize
\begin{tabular}{l?c?c|c?c}
\hline

\hline

\hline
 \bf{Model}      & $D_\text{code}$   & \textit{Valid} ($\downarrow$)~  &  \textit{Test} ($\downarrow$)~~ & $\#$ params ($\downarrow$)  \\ \hline
 
 \hline

\text{Linear Transformer}    & - & 36.473   & 37.533   & \textbf{44.02} $M$     \\

~~~~~~~~~~ {+ \scriptsize{AID}} & - & 36.159  & {37.151}  & 44.16 $M$ \\

\hline

\rowcolor{ABC}
~~~~~~~~~~ {+ \texttt{\textbf{D3}} (\textit{w/o F})} & 32 & \underline{36.061}  & 37.220  & \underline{44.12} $M$\\

\rowcolor{ABC}
 & 64 & \bf{35.975} & \bf{37.009} & 44.36 $M$\\

\hline

\rowcolor{ABC}
~~~~~~~~~~ {+ \texttt{\textbf{D3}} (\textit{w/ F})} & 32 & 36.630  & 37.620  & 44.22 $M$\\

\rowcolor{ABC}
 & 64 & 36.220  & \underline{37.128} & 44.62 $M$\\

 \hline
 
\hline

\hline
\end{tabular}}
\end{center}

\endgroup

\end{minipage}
\vskip -0.1in
\end{table}

Next, we test TPR-RNN and FWM with \texttt{D3} on the sys-bAbI task. This task involves compositional information in each story sentence, such as the relation between objects and their locations. It makes a sentence-level model more suitable for capturing the structural information of data than a word-level model. However, as shown in Table~\ref{table.babi}, TPR-RNN shows a larger performance gap between the \textit{w/o sys diff} and \textit{w/ sys diff} cases than FWM. Notably, \texttt{D3} enhances the systematic generalization of both TPR-RNN and FWM with fewer additional parameters, significantly reducing the performance gap for TPR-RNN. These results highlight the efficacy of \texttt{D3} in text understanding tasks.

\subsubsection{Linear Transformer}

We also evaluate the Linear Transformer with \texttt{D3} on the sort-of-CLEVR task and WikiText-103 task. Following the AID \cite{park2023attention}, we use a 4-layered Linear Transformer with shared parameters for the sort-of-CLEVR task and apply \texttt{D3} to a 16-layered Linear Transformer at intervals of 4 out of the 16 layers for the WikiText-103 task. As shown in Tables~\ref{table.clevr} and~\ref{table.wiki}, \texttt{D3} improves the performance of the Linear Transformer, with these improvements increasing as the capacity of the dictionaries grows. These results demonstrate the effectiveness of \texttt{D3} on visual relational reasoning and language modeling tasks, as well as its applicability to the Linear Transformer. In addition, \texttt{D3} shows comparable performance to the attention-based decomposition method, even with fewer parameters.

\begin{figure}[t]
\begin{center}
\begin{minipage}[t]{0.9\linewidth}
\includegraphics[width=\linewidth]{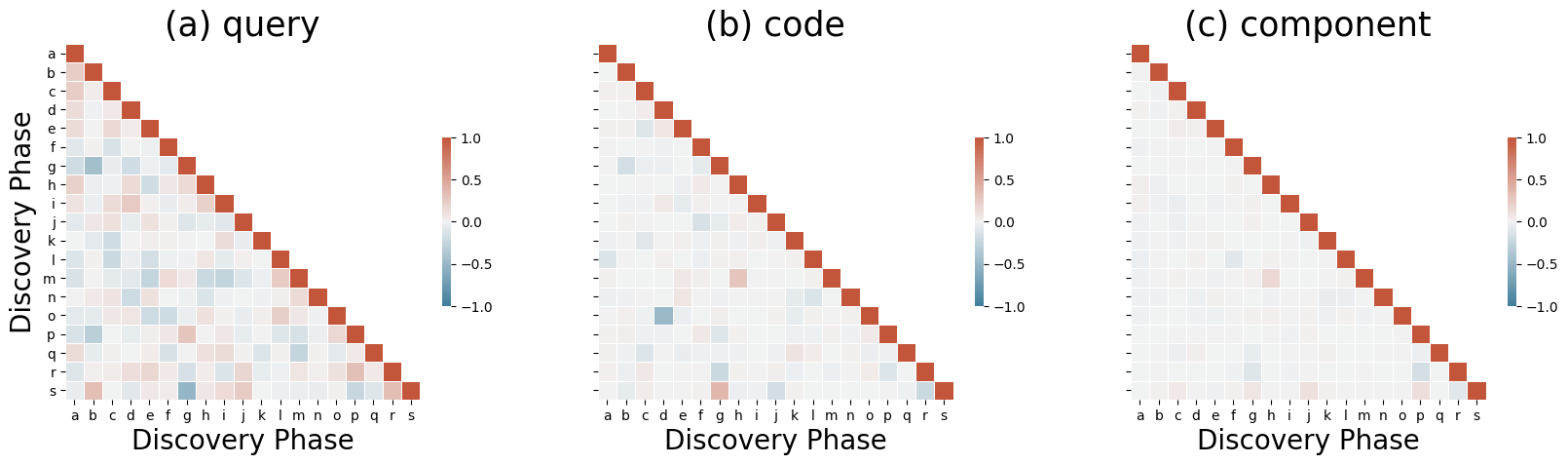}
\caption{
The heatmap displays the cosine similarity between the generated representations during the discovery phase for the SAR task. We explore the similarity across different types of representations: (a) \texttt{queries} of \textit{roles}, (b) \texttt{codes} of \textit{roles}, and (c) the \textit{roles} themselves.
}
\label{fig.sar.analysis.role}
\end{minipage}
\vskip 0.1in
\begin{minipage}[t]{0.9\linewidth}
\includegraphics[width=\linewidth]{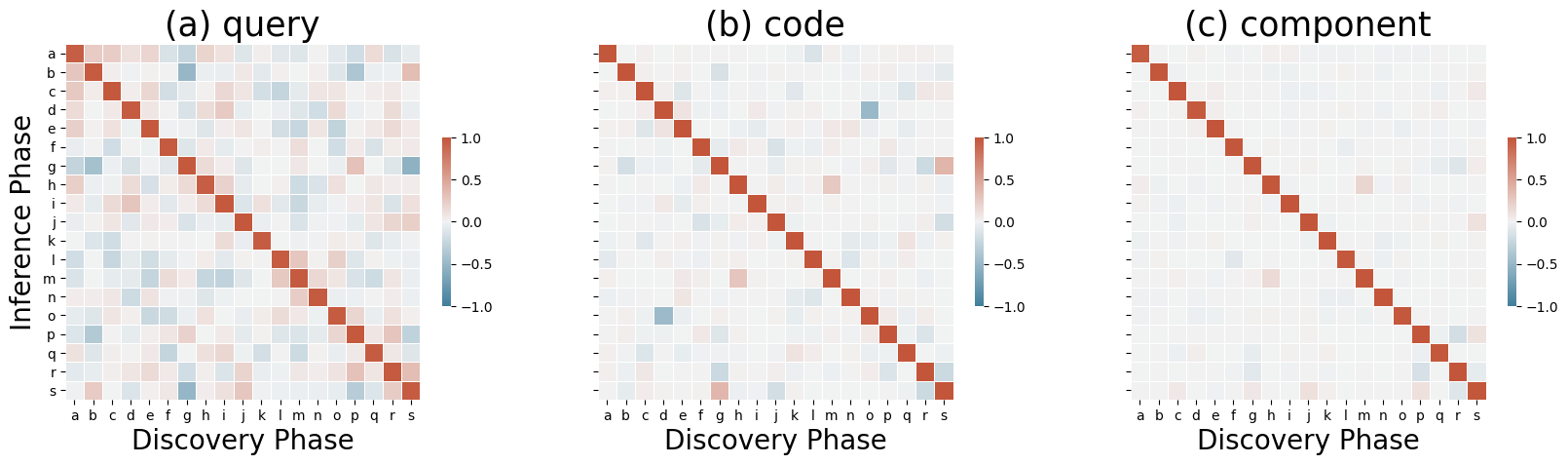}
\caption{
The heatmap displays the cosine similarity between the generated representations during the discovery phase (represented on the \textbf{x-axis}) and the inference phase (represented on the \textbf{y-axis}) for the SAR task. We explore the similarity across different types of representations: (a) \texttt{queries} of \textit{roles} and \textit{unbinding operators}, (b) \texttt{codes} of \textit{roles} and \textit{unbinding operators}, and (c) the \textit{roles} and \textit{unbinding operators} themselves.
}
\label{fig.sar.analysis.role.unbind}
\end{minipage}
\vskip 0.1in
\begin{minipage}[t]{\linewidth}
\begin{center}
\includegraphics[width=0.9\columnwidth]{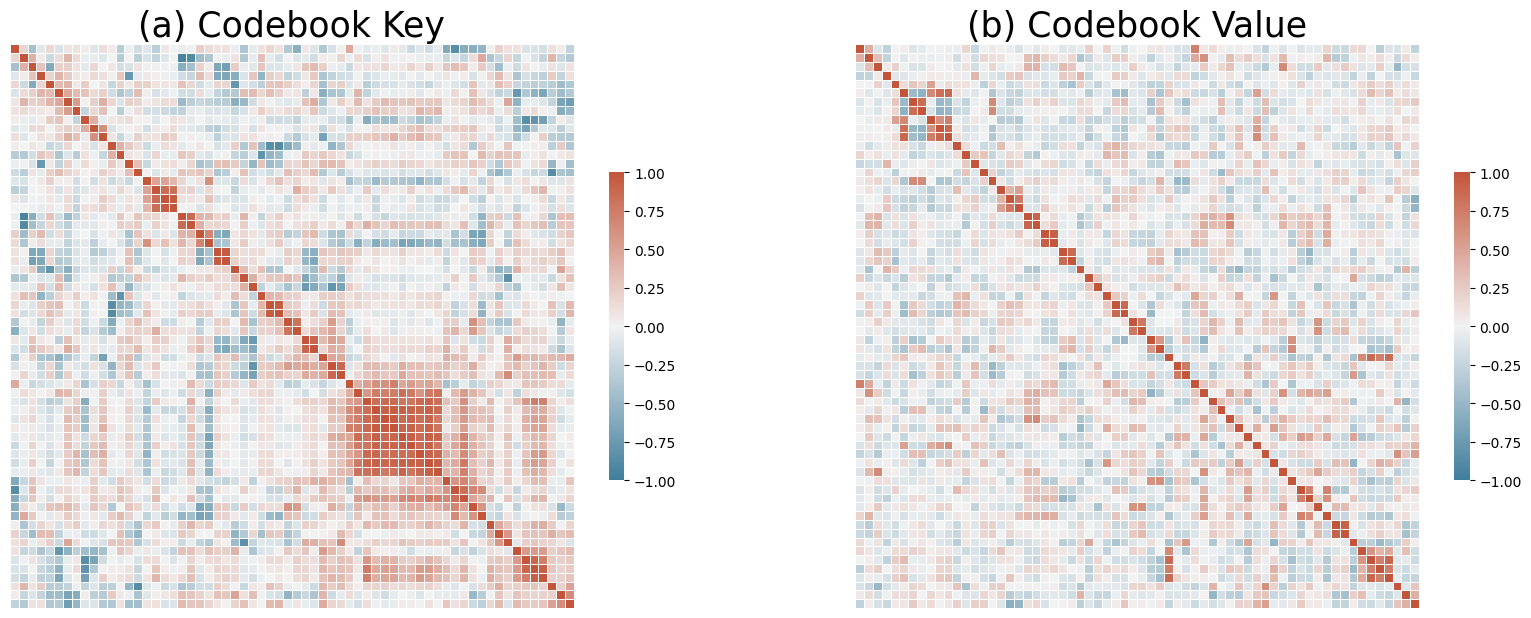}
\caption{
The heatmap visualizes the cosine similarity of the learned codebook features for the SAR task. There are two parts to each heatmap: (a) the similarity among codebook keys, denoted as $\{\textsf{k}_i\}_{i=1}^{N\text{code}}$, and (b) the similarity among codebook values, denoted as $\{\textsf{v}_i\}_{i=1}^{N\text{code}}$. For better visualization, the heatmap values are reordered to reflect the cluster of similar codebook keys.}
\label{fig.kv}
\end{center}
\end{minipage}
\end{center}
\vskip -0.2in
\end{figure}

\subsection{Analysis}

In this section, we conduct a qualitative analysis of the structured TPR representations generated by \texttt{D3} and an ablation study of \texttt{D3}. For these analyses, we experiment with \texttt{D3} (\textit{w/o F}) on the SAR task.

\subsubsection{Qualitative Analysis}\label{sec.orthogonality}

TPR framework requires its structured representations to satisfy the following conditions for accurate TPR operations: (\textit{i}) linearly independence between distinct \textit{roles}, and (\textit{ii}) high correlation between \textit{role} and \textit{unbinding operator} for the same symbol $x$. We analyze the orthogonality of generated representations to investigate whether they satisfy these TPR conditions. Specifically, we consider the case of varying $x$ while keeping $y$ fixed for simplicity.

Fig.~\ref{fig.sar.analysis.role}(c) shows the cosine similarity between the \textit{roles} during the discovery phase, and Fig.~\ref{fig.sar.analysis.role.unbind}(c) shows the cosine similarity between the \textit{roles} during the discovery phase and the \textit{unbinding operator} during the inference phase. Both results demonstrate that the generated representations by \texttt{D3} satisfy the TPR conditions, resulting in an accuracy of nearly 100\%. We also conduct the same analysis for intermediate features, particularly \texttt{query} and \texttt{code}. Figs.~\ref{fig.sar.analysis.role} and~\ref{fig.sar.analysis.role.unbind} show that each intermediate representation complements the others to satisfy the TPR condition, indicating the effectiveness of \texttt{D3}.

Furthermore, we analyze the similarity patterns of codebook keys and codebook values. Fig.~\ref{fig.kv} shows that the codebook features learn orthogonal patterns despite being learned without constraints. This result implies that the learnable parameters of dictionaries implicitly capture TPR conditions to ensure accurate TPR operations.

\begin{figure}[t]
\centering
\begin{subfigure}{.32\textwidth}
  \centering
  \includegraphics[width=\linewidth]{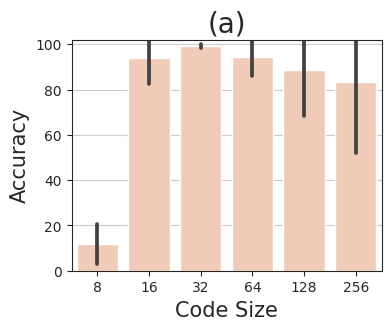}
\end{subfigure}
\begin{subfigure}{.32\textwidth}
  \centering
  \includegraphics[width=\linewidth]{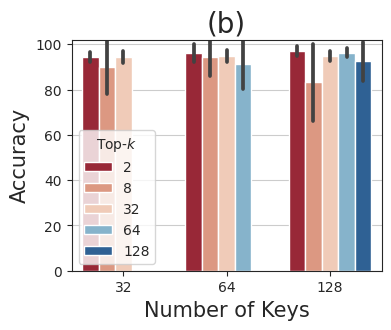}
\end{subfigure}
\begin{subfigure}{.32\textwidth}
  \centering
  \includegraphics[width=\linewidth]{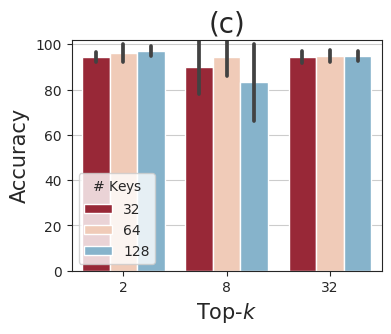}
\end{subfigure}
\caption{
The mean accuracy on the SAR task for 10 seeds in the ablation study, with error bar indicating SD. The default setting uses $D_\text{code}$ of 64, $N_\text{code}$ of 64, and top-$k$ of 8. Each figure shows the experimental results for the following settings: (a) Varying $D_\text{code}$. (b) Varying $N_\text{code}$ with top-$k$ constant. (c) Varying top-$k$ with $N_\text{code}$ constant.
}
\label{fig.sar.ablation}
\vskip -0.1in
\end{figure}

\subsubsection{Ablation Study}

We investigate the effect of hyper-parameters of \texttt{D3}, specifically $N_\text{code}$, $D_\text{code}$, and top-$k$, on performance on the SAR task. Fig.~\ref{fig.sar.ablation}(a) shows the effect of $D_\text{code}$. We observe that the value of $D_\text{code}$ significantly affects the performance of \texttt{D3}. Notably, \texttt{D3} fails to solve the SAR task when $D_\text{code}$ is set to 8, indicating a need for adequate capacity of $D_\text{code}$. Fig.~\ref{fig.sar.ablation}(b) shows the effect of varying top-$k$ while holding $N_\text{code}$ constant, indicating that \texttt{D3} achieves optimal performance when top-$k$ is set to 2. This result demonstrates the efficacy of the sparse mechanism employed by \texttt{D3}. Fig.~\ref{fig.sar.ablation}(c) examines the effect of varying $N_\text{code}$ while holding top-$k$ constant, showing that \texttt{D3} generally performs better with larger values of $N_\text{code}$.

\section{Discussion and Limitations}
\label{sec.discussion}

\paragraph{Motivation.}
From the perspective of systematic generalization, the decomposition operations in the TPR framework can be viewed as mapping unseen data to TPR components observed during training. Motivated by this, we design a decomposition module based on discrete representations, which maps input data to discrete, learned features facilitating systematic generalization in the decomposition operations of TPR. This design choice differentiates our contribution from AID's competitive attention-based decomposition module. Additionally, each dictionary in \texttt{D3} is explicitly linked to a specific TPR component, ensuring that each dictionary is responsible solely for generating its corresponding component. The generated components are then utilized in predefined TPR operations of the TPR-based models. This design ensures that each dictionary is trained to specialize in a specific TPR component.

\paragraph{Interpretability.}
The TPR framework decomposes data at the representation level into distinct symbols, such as \textit{role-filler} pairs for encoding and \textit{unbinding operators} for decoding. This characteristic enhances the interpretability of models because the relationships between \textit{roles} and \textit{unbinding operators} explain which parts of the input the model focuses on to predict the output. However, this interpretability is reliable only when the generated structured representations satisfy the TPR conditions. In this context, \texttt{D3} enhances the interpretability of models by providing structured representations that more effectively satisfy the TPR conditions than baseline models like FWM and AID. Figs.~\ref{fig.sar.baseline.analysis.role} and~\ref{fig.sar.baseline.analysis.role.unbind} demonstrate that the representations generated by \texttt{D3} better conform to the TPR conditions than those from other baseline models, supporting our claim that \texttt{D3} contributes to increased interpretability.

\paragraph{\texttt{D3} Applied to Filler (w/o F and w/ F).}
In the TPR framework, \textit{roles} and \textit{unbinding operators} must meet specific conditions, such as linear independence among \textit{roles} and high correlation between \textit{roles} and \textit{unbinding operators}, to ensure accurate TPR operations. However, there are no such requirements for \textit{fillers}, which are features related to downstream tasks. This characteristic affects the performance of \texttt{D3} depending on whether it is applied to generate the \textit{fillers} (\textit{w/ F}) or not (\textit{w/o F}). In our experiments, the \textit{w/ F} configuration performs well on the sys-bAbI and sort-of-CLEVR tasks with relatively few labels (\textasciitilde 200). In contrast, the \textit{w/o F} configuration excels on the SAR and WikiText-103 tasks, which have a larger number of labels (500\textasciitilde). These findings suggest that the \textit{w/o F} configuration may be more effective for large-scale practical tasks. Nevertheless, beyond these experimental results, we do not fully understand the conditions under which each configuration performs better. Consequently, one limitation of \texttt{D3} is the additional burden of determining the suitable configuration for various tasks when applying it to other domains.

\paragraph{Sparse Key Selection.}
\texttt{D3} integrates seamlessly with existing TPR-based models, significantly enhancing their generalization performance across various tasks. However, this integration introduces additional computational overhead to the baseline models. Specifically, the sparse key selection mechanism of \texttt{D3} has a computational complexity of $\mathcal{O}(N_\text{code} \times (D_\text{query} + \text{log}k))$ for each TPR component. Therefore, this complexity can become a drawback as the capacity of the dictionaries increases. One potential solution to address this capacity issue is to incorporate product keys into the sparse key selection mechanism of \texttt{D3}, a technique studied in prior discrete key-value architectures \cite{lample2019large}. We leave this enhancement for future work.

\paragraph{Scalability.}
The scalability of \texttt{D3} is inherently linked to TPR operations of baseline models since the number of dictionaries in the \texttt{D3} layer aligns with the number of TPR components required for their operations. As TPR operations require increasing components to handle large datasets, our method also requires a proportional increase in dictionaries, resulting in significant computational and memory overhead. As explored in prior work, one potential solution to mitigate this issue is distributing shared dictionaries across multiple heads or layers \cite{lample2019large}. However, this approach requires further investigation and experimentation, which we plan to research in future work.

\section{Conclusion}

In this paper, we tackle the decomposition problem inherent in the TPR framework, which poses a significant challenge for TPR-based models. To address this, we introduce a discrete dictionary-based layer, \texttt{D3}, designed to enhance the decomposition capabilities of TPR-based models. \texttt{D3} employs the discrete dictionaries to map input data to pre-learned symbolic features within each dictionary, thereby generating structured TPR representations. Our comprehensive experiments demonstrate that \texttt{D3} significantly enhances the systematic generalization of the TPR-based models with fewer additional parameters. Furthermore, our qualitative analysis verifies that \texttt{D3} effectively generates structured representations that are satisfactory for the requirements of the TPR framework.

\section*{Acknowledgements}
This work was supported by the National Research Foundation of Korea (NRF) grant funded by the Korea government (MSIT) (No. 2021R1A2C3011169 \& No. 2022R1A5A7026673 \& No. RS-2022-00166735 \& No. RS-2023-00218987).
%This work was supported by the National Research Foundation of Korea(NRF) grant funded by the Korea government(MSIT) (No. 2021R1A2C3011169 \& No. 2022R1A5A7026673) and the National Research Foundation of Korea(NRF) grant funded by the Korea government(MSIT)(No. 2022R1A5A7026673).

%%%%%%%%%%%%%%%%%%%%%%%%%%%%%%%%%%%%%%%%%%%%%%%%%%%%%%%%%%%%
\bibliography{reference}

\begin{thebibliography}{44}
\providecommand{\natexlab}[1]{#1}
\providecommand{\url}[1]{\texttt{#1}}
\expandafter\ifx\csname urlstyle\endcsname\relax
  \providecommand{\doi}[1]{doi: #1}\else
  \providecommand{\doi}{doi: \begingroup \urlstyle{rm}\Url}\fi

\bibitem[Ba et~al.(2016)Ba, Kiros, and Hinton]{ba2016layer}
J.~L. Ba, J.~R. Kiros, and G.~E. Hinton.
\newblock Layer normalization.
\newblock \emph{arXiv preprint arXiv:1607.06450}, 2016.

\bibitem[Fodor and Pylyshyn(1988)]{fodor1988connectionism}
J.~A. Fodor and Z.~W. Pylyshyn.
\newblock Connectionism and cognitive architecture: A critical analysis.
\newblock \emph{Cognition}, 28\penalty0 (1-2):\penalty0 3--71, 1988.

\bibitem[Goyal et~al.(2019)Goyal, Lamb, Hoffmann, Sodhani, Levine, Bengio, and Sch{\"o}lkopf]{goyal2019recurrent}
A.~Goyal, A.~Lamb, J.~Hoffmann, S.~Sodhani, S.~Levine, Y.~Bengio, and B.~Sch{\"o}lkopf.
\newblock Recurrent independent mechanisms.
\newblock \emph{arXiv preprint arXiv:1909.10893}, 2019.

\bibitem[Graves et~al.(2014)Graves, Wayne, and Danihelka]{graves2014neural}
A.~Graves, G.~Wayne, and I.~Danihelka.
\newblock Neural turing machines.
\newblock \emph{arXiv preprint arXiv:1410.5401}, 2014.

\bibitem[Graves et~al.(2016)Graves, Wayne, Reynolds, Harley, Danihelka, Grabska-Barwi{\'n}ska, Colmenarejo, Grefenstette, Ramalho, Agapiou, et~al.]{graves2016hybrid}
A.~Graves, G.~Wayne, M.~Reynolds, T.~Harley, I.~Danihelka, A.~Grabska-Barwi{\'n}ska, S.~G. Colmenarejo, E.~Grefenstette, T.~Ramalho, J.~Agapiou, et~al.
\newblock Hybrid computing using a neural network with dynamic external memory.
\newblock \emph{Nature}, 538\penalty0 (7626):\penalty0 471--476, 2016.

\bibitem[Greff et~al.(2020)Greff, Van~Steenkiste, and Schmidhuber]{greff2020binding}
K.~Greff, S.~Van~Steenkiste, and J.~Schmidhuber.
\newblock On the binding problem in artificial neural networks.
\newblock \emph{arXiv preprint arXiv:2012.05208}, 2020.

\bibitem[Hsu et~al.(2024)Hsu, Dorrell, Whittington, Wu, and Finn]{hsu2024disentanglement}
K.~Hsu, W.~Dorrell, J.~Whittington, J.~Wu, and C.~Finn.
\newblock Disentanglement via latent quantization.
\newblock \emph{Advances in Neural Information Processing Systems}, 36, 2024.

\bibitem[Hupkes et~al.(2020)Hupkes, Dankers, Mul, and Bruni]{hupkes2020compositionality}
D.~Hupkes, V.~Dankers, M.~Mul, and E.~Bruni.
\newblock Compositionality decomposed: How do neural networks generalise?
\newblock \emph{Journal of Artificial Intelligence Research}, 67:\penalty0 757--795, 2020.

\bibitem[Jiang et~al.(2021)Jiang, Celikyilmaz, Smolensky, Soulos, Rao, Palangi, Fernandez, Smith, Bansal, and Gao]{jiang2021enriching}
Y.~Jiang, A.~Celikyilmaz, P.~Smolensky, P.~Soulos, S.~Rao, H.~Palangi, R.~Fernandez, C.~Smith, M.~Bansal, and J.~Gao.
\newblock Enriching transformers with structured tensor-product representations for abstractive summarization.
\newblock \emph{arXiv preprint arXiv:2106.01317}, 2021.

\bibitem[Katharopoulos et~al.(2020)Katharopoulos, Vyas, Pappas, and Fleuret]{katharopoulos2020transformers}
A.~Katharopoulos, A.~Vyas, N.~Pappas, and F.~Fleuret.
\newblock Transformers are rnns: Fast autoregressive transformers with linear attention.
\newblock In \emph{International conference on machine learning}, pages 5156--5165. PMLR, 2020.

\bibitem[Kori et~al.(2023)Kori, Locatello, Ribeiro, Toni, and Glocker]{kori2023grounded}
A.~Kori, F.~Locatello, F.~D.~S. Ribeiro, F.~Toni, and B.~Glocker.
\newblock Grounded object-centric learning.
\newblock In \emph{The Twelfth International Conference on Learning Representations}, 2023.

\bibitem[Lake and Baroni(2018)]{lake2018generalization}
B.~Lake and M.~Baroni.
\newblock Generalization without systematicity: On the compositional skills of sequence-to-sequence recurrent networks.
\newblock In \emph{International conference on machine learning}, pages 2873--2882. PMLR, 2018.

\bibitem[Lake et~al.(2017)Lake, Ullman, Tenenbaum, and Gershman]{lake2017building}
B.~M. Lake, T.~D. Ullman, J.~B. Tenenbaum, and S.~J. Gershman.
\newblock Building machines that learn and think like people.
\newblock \emph{Behavioral and brain sciences}, 40:\penalty0 e253, 2017.

\bibitem[Lample et~al.(2019)Lample, Sablayrolles, Ranzato, Denoyer, and J{\'e}gou]{lample2019large}
G.~Lample, A.~Sablayrolles, M.~Ranzato, L.~Denoyer, and H.~J{\'e}gou.
\newblock Large memory layers with product keys.
\newblock \emph{Advances in Neural Information Processing Systems}, 32, 2019.

\bibitem[Le et~al.(2020)Le, Tran, and Venkatesh]{le2020self}
H.~Le, T.~Tran, and S.~Venkatesh.
\newblock Self-attentive associative memory.
\newblock In \emph{International Conference on Machine Learning}, pages 5682--5691. PMLR, 2020.

\bibitem[Li{\v{s}}ka et~al.(2018)Li{\v{s}}ka, Kruszewski, and Baroni]{livska2018memorize}
A.~Li{\v{s}}ka, G.~Kruszewski, and M.~Baroni.
\newblock Memorize or generalize? searching for a compositional rnn in a haystack.
\newblock \emph{arXiv preprint arXiv:1802.06467}, 2018.

\bibitem[Liu et~al.(2021)Liu, Lamb, Kawaguchi, ALIAS PARTH~GOYAL, Sun, Mozer, and Bengio]{liu2021discrete}
D.~Liu, A.~M. Lamb, K.~Kawaguchi, A.~G. ALIAS PARTH~GOYAL, C.~Sun, M.~C. Mozer, and Y.~Bengio.
\newblock Discrete-valued neural communication.
\newblock \emph{Advances in Neural Information Processing Systems}, 34:\penalty0 2109--2121, 2021.

\bibitem[Locatello et~al.(2020)Locatello, Weissenborn, Unterthiner, Mahendran, Heigold, Uszkoreit, Dosovitskiy, and Kipf]{locatello2020object}
F.~Locatello, D.~Weissenborn, T.~Unterthiner, A.~Mahendran, G.~Heigold, J.~Uszkoreit, A.~Dosovitskiy, and T.~Kipf.
\newblock Object-centric learning with slot attention.
\newblock \emph{Advances in Neural Information Processing Systems}, 33:\penalty0 11525--11538, 2020.

\bibitem[Merity et~al.(2016)Merity, Xiong, Bradbury, and Socher]{merity2016pointer}
S.~Merity, C.~Xiong, J.~Bradbury, and R.~Socher.
\newblock Pointer sentinel mixture models.
\newblock \emph{arXiv preprint arXiv:1609.07843}, 2016.

\bibitem[Mittal et~al.(2021)Mittal, Raparthy, Rish, Bengio, and Lajoie]{mittal2021compositional}
S.~Mittal, S.~C. Raparthy, I.~Rish, Y.~Bengio, and G.~Lajoie.
\newblock Compositional attention: Disentangling search and retrieval.
\newblock \emph{arXiv preprint arXiv:2110.09419}, 2021.

\bibitem[Palangi et~al.(2018)Palangi, Smolensky, He, and Deng]{palangi2018question}
H.~Palangi, P.~Smolensky, X.~He, and L.~Deng.
\newblock Question-answering with grammatically-interpretable representations.
\newblock In \emph{Proceedings of the AAAI Conference on Artificial Intelligence}, volume~32, 2018.

\bibitem[Park et~al.(2021)Park, Choi, and Lee]{park2021distributed}
T.~Park, I.~Choi, and M.~Lee.
\newblock Distributed associative memory network with memory refreshing loss.
\newblock \emph{Neural Networks}, 144:\penalty0 33--48, 2021.

\bibitem[Park et~al.(2023)Park, Choi, and Lee]{park2023attention}
T.~Park, I.~Choi, and M.~Lee.
\newblock Attention-based iterative decomposition for tensor product representation.
\newblock In \emph{The Twelfth International Conference on Learning Representations}, 2023.

\bibitem[Rae et~al.(2016)Rae, Hunt, Danihelka, Harley, Senior, Wayne, Graves, and Lillicrap]{rae2016scaling}
J.~Rae, J.~J. Hunt, I.~Danihelka, T.~Harley, A.~W. Senior, G.~Wayne, A.~Graves, and T.~Lillicrap.
\newblock Scaling memory-augmented neural networks with sparse reads and writes.
\newblock In \emph{Advances in Neural Information Processing Systems}, pages 3621--3629, 2016.

\bibitem[Rosenblatt(1958)]{rosenblatt1958perceptron}
F.~Rosenblatt.
\newblock The perceptron: a probabilistic model for information storage and organization in the brain.
\newblock \emph{Psychological review}, 65\penalty0 (6):\penalty0 386, 1958.

\bibitem[Santoro et~al.(2017)Santoro, Raposo, Barrett, Malinowski, Pascanu, Battaglia, and Lillicrap]{santoro2017simple}
A.~Santoro, D.~Raposo, D.~G. Barrett, M.~Malinowski, R.~Pascanu, P.~Battaglia, and T.~Lillicrap.
\newblock A simple neural network module for relational reasoning.
\newblock \emph{Advances in neural information processing systems}, 30, 2017.

\bibitem[Santoro et~al.(2018)Santoro, Faulkner, Raposo, Rae, Chrzanowski, Weber, Wierstra, Vinyals, Pascanu, and Lillicrap]{santoro2018relational}
A.~Santoro, R.~Faulkner, D.~Raposo, J.~Rae, M.~Chrzanowski, T.~Weber, D.~Wierstra, O.~Vinyals, R.~Pascanu, and T.~Lillicrap.
\newblock Relational recurrent neural networks.
\newblock \emph{Advances in neural information processing systems}, 31, 2018.

\bibitem[Schlag and Schmidhuber(2018)]{schlag2018learning}
I.~Schlag and J.~Schmidhuber.
\newblock Learning to reason with third order tensor products.
\newblock \emph{Advances in neural information processing systems}, 31, 2018.

\bibitem[Schlag et~al.(2019)Schlag, Smolensky, Fernandez, Jojic, Schmidhuber, and Gao]{schlag2019enhancing}
I.~Schlag, P.~Smolensky, R.~Fernandez, N.~Jojic, J.~Schmidhuber, and J.~Gao.
\newblock Enhancing the transformer with explicit relational encoding for math problem solving.
\newblock \emph{arXiv preprint arXiv:1910.06611}, 2019.

\bibitem[Schlag et~al.(2020)Schlag, Munkhdalai, and Schmidhuber]{schlag2020learning}
I.~Schlag, T.~Munkhdalai, and J.~Schmidhuber.
\newblock Learning associative inference using fast weight memory.
\newblock \emph{arXiv preprint arXiv:2011.07831}, 2020.

\bibitem[Schlag et~al.(2021)Schlag, Irie, and Schmidhuber]{schlag2021linear}
I.~Schlag, K.~Irie, and J.~Schmidhuber.
\newblock Linear transformers are secretly fast weight programmers.
\newblock In \emph{International Conference on Machine Learning}, pages 9355--9366. PMLR, 2021.

\bibitem[Shi et~al.(2022)Shi, Zhang, and Lipani]{shi2022stepgame}
Z.~Shi, Q.~Zhang, and A.~Lipani.
\newblock Stepgame: A new benchmark for robust multi-hop spatial reasoning in texts.
\newblock In \emph{Proceedings of the AAAI conference on artificial intelligence}, volume~36, pages 11321--11329, 2022.

\bibitem[Smolensky(1990)]{smolensky1990tensor}
P.~Smolensky.
\newblock Tensor product variable binding and the representation of symbolic structures in connectionist systems.
\newblock \emph{Artificial intelligence}, 46\penalty0 (1-2):\penalty0 159--216, 1990.

\bibitem[Soulos et~al.(2023)Soulos, Hu, McCurdy, Chen, Fernandez, Smolensky, and Gao]{soulos2023differentiable}
P.~Soulos, E.~J. Hu, K.~McCurdy, Y.~Chen, R.~Fernandez, P.~Smolensky, and J.~Gao.
\newblock Differentiable tree operations promote compositional generalization.
\newblock In \emph{International Conference on Machine Learning}, pages 32499--32520. PMLR, 2023.

\bibitem[Srivastava et~al.(2014)Srivastava, Hinton, Krizhevsky, Sutskever, and Salakhutdinov]{srivastava2014dropout}
N.~Srivastava, G.~Hinton, A.~Krizhevsky, I.~Sutskever, and R.~Salakhutdinov.
\newblock Dropout: a simple way to prevent neural networks from overfitting.
\newblock \emph{The journal of machine learning research}, 15\penalty0 (1):\penalty0 1929--1958, 2014.

\bibitem[Sukhbaatar et~al.(2015)Sukhbaatar, Weston, Fergus, et~al.]{sukhbaatar2015end}
S.~Sukhbaatar, J.~Weston, R.~Fergus, et~al.
\newblock End-to-end memory networks.
\newblock In \emph{Advances in neural information processing systems}, pages 2440--2448, 2015.

\bibitem[Tamkin et~al.(2023)Tamkin, Taufeeque, and Goodman]{tamkin2023codebook}
A.~Tamkin, M.~Taufeeque, and N.~D. Goodman.
\newblock Codebook features: Sparse and discrete interpretability for neural networks.
\newblock \emph{arXiv preprint arXiv:2310.17230}, 2023.

\bibitem[Tr{\"a}uble et~al.(2023)Tr{\"a}uble, Goyal, Rahaman, Mozer, Kawaguchi, Bengio, and Sch{\"o}lkopf]{trauble2023discrete}
F.~Tr{\"a}uble, A.~Goyal, N.~Rahaman, M.~C. Mozer, K.~Kawaguchi, Y.~Bengio, and B.~Sch{\"o}lkopf.
\newblock Discrete key-value bottleneck.
\newblock In \emph{International Conference on Machine Learning}, pages 34431--34455. PMLR, 2023.

\bibitem[Van Den~Oord et~al.(2017)Van Den~Oord, Vinyals, et~al.]{van2017neural}
A.~Van Den~Oord, O.~Vinyals, et~al.
\newblock Neural discrete representation learning.
\newblock \emph{Advances in neural information processing systems}, 30, 2017.

\bibitem[Vaswani et~al.(2017)Vaswani, Shazeer, Parmar, Uszkoreit, Jones, Gomez, Kaiser, and Polosukhin]{vaswani2017attention}
A.~Vaswani, N.~Shazeer, N.~Parmar, J.~Uszkoreit, L.~Jones, A.~N. Gomez, {\L}.~Kaiser, and I.~Polosukhin.
\newblock Attention is all you need.
\newblock \emph{Advances in neural information processing systems}, 30, 2017.

\bibitem[Webb et~al.(2020)Webb, Sinha, and Cohen]{webb2020emergent}
T.~W. Webb, I.~Sinha, and J.~D. Cohen.
\newblock Emergent symbols through binding in external memory.
\newblock \emph{arXiv preprint arXiv:2012.14601}, 2020.

\bibitem[Weston et~al.(2015)Weston, Bordes, Chopra, Rush, Van~Merri{\"e}nboer, Joulin, and Mikolov]{weston2015towards}
J.~Weston, A.~Bordes, S.~Chopra, A.~M. Rush, B.~Van~Merri{\"e}nboer, A.~Joulin, and T.~Mikolov.
\newblock Towards ai-complete question answering: A set of prerequisite toy tasks.
\newblock \emph{arXiv preprint arXiv:1502.05698}, 2015.

\bibitem[Wu et~al.(2024)Wu, Lee, and Ahn]{wu2024structured}
Y.-F. Wu, M.~Lee, and S.~Ahn.
\newblock Structured world modeling via semantic vector quantization.
\newblock \emph{arXiv preprint arXiv:2402.01203}, 2024.

\bibitem[Zhuang et~al.(2024)Zhuang, Zhang, Ding, Bian, Wang, Lv, Chen, and Chen]{zhuang2024learning}
X.~Zhuang, Q.~Zhang, K.~Ding, Y.~Bian, X.~Wang, J.~Lv, H.~Chen, and H.~Chen.
\newblock Learning invariant molecular representation in latent discrete space.
\newblock \emph{Advances in Neural Information Processing Systems}, 36, 2024.

\end{thebibliography}

\newpage
\begin{appendices}

\section{Experiment Details}
\label{appendix.experiment.design}

This section provides a detailed description of our experiments on the SAR task, sys-bAbI task, sort-of-CLEVR task, and WikiText-103 task. We followed the experimental settings outlined by AID \cite{park2023attention} to assess the decomposition capabilities of \texttt{D3}. To ensure stability and reproducibility, we ran all experiments, except for the WikiText-103 task, using 10 different random seeds\footnote{We used the following seed values: \{0, 1111, 2222, 3333, 4444, 5555, 6666, 7777, 8888, 9999\}}. For the WikiText-103 task, we experimented with a single seed of 1111. Each experiment was conducted on a single 48GB NVIDIA RTX A6000 GPU and an AMD EPYC 7513 32-Core Processor.

% \footnote{The code of AID is publicly available at \href{https://github.com/taewonpark/AID}{https://github.com/taewonpark/AID}}

\subsection{Systematic Associative Recall task}

The SAR task \cite{park2023attention} evaluates systematic generalization in memorizing and recalling combinatorial data. It consists of a discovery phase and an inference phase. During the discovery phase, the model receives the combinatorial sequential items, each combining two symbols, $x \in X$ and $y \in Y$ where $X=X_1 \cup X_2 \cup X_3$ and $Y=Y_1 \cup Y_2$. The model is then required to predict an associated $y$ when a specific $x$ is presented. The SAR task uses different combination settings between training and evaluation to target systematic generalization specifically. During the training, the model learns the following combination settings: (1) $X_1$ and $Y_1$, (2) $X_2$ and $Y_2$, and (3) $X_3$ and $Y$. At evaluation, however, the model should generalize unseen combination settings, specifically $X_1$ and $Y_2$. In our study, unlike the AID paper \cite{park2023attention}, we only consider the most challenging setting of the SAR task by excluding the subset $X_3$.

Each combinatorial item is constructed as follows. First, symbols $x$ and $y$ are sampled from their respective sets $X$ and $Y$, where $|X_1|=|X_2|=|Y_1|=|Y_2|=250$. The sampled symbols are mapped into a 50-dimensional space using a word embedding method. These embedding vectors are then concatenated to construct the combinatorial item. For training, 100 randomly generated combinatorial items are sequentially provided to the model during the discovery phase. During the inference phase, the model receives only the $x$ symbols sequentially, with the embedding vector of $y$ set to zero. This task also provides binary flags to indicate the start of each phase. At evaluation, all possible combinations that can be formed in $X_1$ and $Y_2$ are tested.

To build the experimental environment for the SAR task, we utilize the open-source implementation\footnote{\href{https://github.com/taewonpark/AID/tree/main/SARtask}{https://github.com/taewonpark/AID/tree/main/SARtask}} from the AID \cite{park2023attention}. We train the model using the Adam optimizer with a batch size of 64 and a learning rate of 1$e^{-3}$, $\beta_1$ of 0.9, and $\beta_2$ of 0.98 for training iterations of 30$K$. Each experiment took approximately 3 hours per each seed.

\subsection{Systematic bAbI task}

The sys-bAbI task \cite{park2023attention} is a variant of the bAbI task \cite{weston2015towards} designed to evaluate systematic generalization in text understanding and reasoning. It consists of 20 distinct sub-tasks, each comprising stories, relevant queries, and corresponding answers. The sys-bAbI task requires the models to remember the stories and predict corresponding answers to the queries. Unlike the original bAbI task, the sys-bAbI task evaluates the models with two aspects: (a) in-distribution (\textit{w/o sys diff}) and (b) with the systematic difference (\textit{w/ sys diff}) where each sub-task includes unseen words during training. Therefore, the models should learn task-independent text understanding to solve the sys-bAbI task.

The bAbI dataset includes various versions, such as \texttt{en-10k} and \texttt{en-valid-10k}. The sys-bAbI task uses the \texttt{en-valid-10k} version, which is already divided into training, validation, and test datasets. To create the experimental environment for the sys-bAbI task, we use the open-source implementation\footnote{\href{https://github.com/taewonpark/AID/tree/main/bAbItask}{https://github.com/taewonpark/AID/tree/main/bAbItask}} provided by the AID.

We use the open-source implementation of the baseline models, TPR-RNN\footnote{\href{https://github.com/APodolskiy/TPR-RNN-Torch}{https://github.com/APodolskiy/TPR-RNN-Torch}} \citep{schlag2018learning} and FWM\footnote{\href{https://github.com/ischlag/Fast-Weight-Memory-public}{https://github.com/ischlag/Fast-Weight-Memory-public}} \citep{schlag2020learning}. Following the experimental settings of baseline models, we use different configurations for each model. We train the TPR-RNN with \texttt{D3} using an embedding size of 179 and the Adam optimizer with a batch size of 128 and a learning rate of 1$e^{-3}$, $\beta_1$ of 0.9, and $\beta_2$ of 0.99 for 100 training epochs. For FWM with \texttt{D3}, we use an embedding size of 256 and the Adam optimizer with a batch size of 64 and a learning rate of 1$e^{-3}$, $\beta_1$ of 0.9, and $\beta_2$ of 0.98 for training iterations of 60$K$. Furthermore, following the AID, we use the reconstruction loss for the bAbI task, introduced in \citet{park2021distributed}, in our experiments on the sys-bAbI task. Each experiment took approximately 7 hours per seed for the TPR-RNN with \texttt{D3} and 8 hours per seed for the FWM with \texttt{D3}.

\subsection{Sort-of-CLEVR task}

The sort-of-CLEVR task \cite{santoro2017simple} evaluates compositional generalization in visual relational reasoning. It consists of scene images, queries, and corresponding answers. This task requires the models to understand the properties of individual objects (\textit{Unary}) or the relationships between multiple objects (\textit{Binary} or \textit{Ternary}) within visual scene images and predict the correct answers to the queries. Therefore, the model should capture relationships within each object and between objects to solve this task.

Each scene image, with a size of 75$\times$75 pixels, includes 6 distinct objects in 6 different colors (red, blue, green, orange, yellow, or gray) and 2 different shapes (square or circle). This scene image is encoded by a visual encoder. The encoded visual feature is then concatenated with the embedding vector of the query. These concatenated features are provided to the model. Following the experimental settings of the AID \cite{park2023attention}, we use a single CNN layer with a kernel size of 15 and a stride of 15 for the visual encoder, and an embedding size of 128 for the word embedding method. Also, we use a 4-layered Transformer, where each layer shares its parameters with others, as our baseline model.

To build the experimental environment for the sort of CLEVR task, we utilize the open-source implementation\footnote{\href{https://github.com/sarthmit/Compositional-Attention/tree/main/Sort-of-CLEVR}{https://github.com/sarthmit/Compositional-Attention/tree/main/Sort-of-CLEVR}} from \citet{mittal2021compositional}. We train the model using the Adam optimizer with a batch size of 64 and a learning rate of 1$e^{-4}$ for 100 training epochs. Each experiment took approximately 2.5 hours per each seed.

\subsection{WikiText-103 task}

The WikiText-103 task \cite{merity2016pointer} is a language modeling dataset consisting of lengthy corpora from Wikipedia. Although the WikiText-103 task does not directly measure the systematic generalization of the models, it is used to evaluate the effectiveness and applicability of \texttt{D3} on a large-scale task beyond relatively simple tasks.

The WikiText-103 task comprises 28,475 articles for training, 60 for validation, and 60 for testing. Following the experimental settings of \citet{schlag2021linear}, we partition the articles into segments of $L$ words. During training, the gradient is back-propagated only within spans of $L$ words. The performance of the model is evaluated using the measure of perplexity. During evaluation, the model processes an input sequence of $L$ words by sliding a segment over the article with a stride size of 1. Perplexity is then computed based on the last position of each segment, except for the first segment, where every position is taken into account.

To build the experimental environment for the WikiText-103 task, we utilize the open-source implementation\footnote{\href{https://github.com/IDSIA/lmtool-fwp}{https://github.com/IDSIA/lmtool-fwp}} from \cite{schlag2021linear}. Following the AID \cite{park2023attention}, we apply \texttt{D3} to a 16-layered Linear Transformer at intervals of 4 out of the 16 layers. We train the model using the Adam optimizer with a batch size of 96, an initial learning rate of 2.5$e^{-4}$, and a learning rate warmup step of 2,000 for 120 epochs. Each experiment took approximately \textasciitilde3 days.

\newpage

\section{Hyper-parameter Settings}
\label{appendix.experiment.parameter}

\begingroup
\setlength{\tabcolsep}{6pt} % Default value: 6pt
\renewcommand{\arraystretch}{1.5} % Default value: 1
\begin{table}[h]
\caption{Hyper-parameter settings of the \texttt{D3}.}
\label{table.parameter.d3}
\begin{center}
{\footnotesize
\begin{tabular}{l?c|c|c|c}
\hline

\hline

\hline
 & \bf SAR task & \bf \textit{sys-bAbI} task & \bf Sort-of-CLEVR task & \bf WikiText-103 task  \\
 \hline

 \hline
 $D_\text{code}$ & 8, 16, \underline{32}, 64, 128 &  32, \underline{64}, 128, 256  & 128, \underline{256} & 32, \underline{64} \\
 \cline{2-5} 
 $N_\text{code}$ & \multicolumn{4}{c}{64}  \\
 $D_\text{query}$ & \multicolumn{4}{c}{$D_\text{code}/2$}  \\
 top-$k$ & \multicolumn{4}{c}{8}  \\
 $p_\text{dropout}$  & \multicolumn{4}{c}{0.1}  \\
\hline

\hline

\hline
\end{tabular}}
\end{center}
\end{table}
\endgroup

\begin{table}[!htb]
\centering
\begin{minipage}[t]{\textwidth}
\centering
    \caption{\label{table.parameter.tpr_rnn}Hyper-parameters of TPR-RNN.}

\begingroup
\setlength{\tabcolsep}{6pt} % Default value: 6pt
\renewcommand{\arraystretch}{1.5} % Default value: 1
\begin{center}
{\footnotesize
\begin{tabular}{l?c}
\hline

\hline

\hline
 &  \bf \textit{sys-bAbI} task \\
 \hline

 \hline

 $D_\text{entity}$ ($D_\text{component}$)  & 90  \\
 $D_\text{relation}$  ($D_\text{component}$)  & 20  \\
 $N_\text{component}^\text{enc}$  & 5  \\
 $N_\text{component}^\text{dec}$  & 4  \\
\hline

\hline

\hline
\end{tabular}}
\end{center}
\endgroup

\end{minipage}%

\bigskip{}

\begin{minipage}[t]{\textwidth}

\centering
    \caption{\label{table.parameter.fwm}Hyper-parameters of FWM.}

\begingroup
\setlength{\tabcolsep}{6pt} % Default value: 6pt
\renewcommand{\arraystretch}{1.5} % Default value: 1
\begin{center}
{\footnotesize
\begin{tabular}{l?c|c}
\hline

\hline

\hline
 & \bf SAR task & \bf \textit{sys-bAbI} task \\
 \hline

 \hline

 $D_\text{LSTM}$  & 256 & 256  \\
 $D_\text{FWM}$  ($D_\text{component}$)  & 32 & 32  \\
 $N_\text{reads}$  & 1 & 3  \\
 $N_\text{component}^\text{enc}$  & 3 & 3  \\
 $N_\text{component}^\text{dec}$  & 1+$N_\text{reads}$ & 1+$N_\text{reads}$  \\
\hline

\hline

\hline
\end{tabular}}
\end{center}
\endgroup

\end{minipage}

\bigskip{}

\begin{minipage}[t]{\textwidth}

\centering
    \caption{\label{table.parameter.linear}Hyper-parameters of Linear Transformer.}

\begingroup
\setlength{\tabcolsep}{6pt} % Default value: 6pt
\renewcommand{\arraystretch}{1.5} % Default value: 1
\begin{center}
{\footnotesize
\begin{tabular}{l?c|c}
\hline

\hline

\hline
 &  \bf Sort-of-CLEVR task & \bf WikiText-103 task  \\
 \hline

 \hline

 $D_\text{heads}$  ($D_\text{component}$)  & 64 & 16 \\
 $N_\text{heads}$  & 4  & 8 \\
 $N_\text{component}^\text{enc}$  & 2 * $N_\text{heads}$ & 2 * $N_\text{heads}$  \\
 $N_\text{component}^\text{dec}$  & $N_\text{heads}$ & $N_\text{heads}$  \\
\hline

\hline

\hline
\end{tabular}}
\end{center}
\endgroup

\end{minipage}
\end{table}

\newpage

\section{Additional Experiments}

\begingroup
\setlength{\tabcolsep}{4pt} % Default value: 6pt
\renewcommand{\arraystretch}{1.2} % Default value: 1

\begin{table}[h]
\caption{The mean word error rate [\%] on additional experiments of the \textit{sys-bAbI} task for 10 seeds.}
%\vskip -0.1in
\label{table.babi.ablation}
\begin{center}
%{\scriptsize%\footnotesize
{\footnotesize
\begin{tabular}{l?c?l|l?c?c}
\hline

\hline

\hline
\bf Model        &  $D_\text{code}$   & \textit{w/o sys diff} ($\downarrow$)   &  \textit{w/ sys diff} ($\downarrow$)     & \textbf{Gap} ($\downarrow$)  & $\#$ params ($\downarrow$) \\ \hline

\hline

\addlinespace[1ex]

\hline

\hline

TPR-RNN &  - & 0.79 \tiny{$\pm$ 0.16}   & 8.74 \tiny{$\pm$ 3.74}    & 7.95  & \underline{0.14} $M$   \\

~~~~~~{+ \scriptsize{AID}} & - & 0.69 \tiny{$\pm$ 0.08} & 5.61 \tiny{$\pm$ 1.78}  & 4.92  & 0.32 $M$ \\

\hline

\rowcolor{ABC}
~~~~~~{+ \texttt{\textbf{D3}}} & 32 & 1.16 \tiny{$\pm$ 0.25}  & \textbf{3.44} \tiny{$\pm$ 1.78}  & \textbf{2.28} & \textbf{0.13} $M$\\
\rowcolor{ABC}
 & 64 & \textbf{0.65} \tiny{$\pm$ 0.25}  & \underline{3.50} \tiny{$\pm$ 2.07}  & \underline{2.85} & 0.17 $M$\\
\rowcolor{ABC}
 & 128 & \underline{0.68} \tiny{$\pm$ 0.14}  & 3.94 \tiny{$\pm$ 2.20}  & 3.26 & 0.26 $M$\\

\hline

\hline

\addlinespace[1ex]

\hline

\hline

FWM     & -  & 0.79 \tiny{$\pm$ 0.14}   & 2.85 \tiny{$\pm$ 1.61}    & 2.06   & \textbf{0.73} $M$  \\

 ~~~~~~{+ \scriptsize{AID}} & - & \textbf{0.45} \tiny{$\pm$ 0.16}  & \textbf{1.21} \tiny{$\pm$ 0.66}  & \textbf{0.76}   & 1.23 $M$ \\

\hline
 
\rowcolor{ABC}
 ~~~~~~{+ \texttt{\textbf{D3}} (\textit{w/o F})} & 64 & 0.79 \tiny{$\pm$ 0.30}  & 2.58 \tiny{$\pm$ 1.12}  & 1.79  & 0.75 $M$ \\

 \rowcolor{ABC}
 & 128 & 0.93 \tiny{$\pm$ 0.20}  & 3.82 \tiny{$\pm$ 1.21}  & 2.89  & 0.82 $M$ \\

 \rowcolor{ABC}
 & 256 & 1.04 \tiny{$\pm$ 0.40}  & 3.33 \tiny{$\pm$ 1.21}  & 2.29  & 0.97 $M$ \\

\hline

\rowcolor{ABC}
 ~~~~~~{+ \texttt{\textbf{D3}} (\textit{w/ F})} & 32 & 1.20 \tiny{$\pm$ 0.31}  & 7.23 \tiny{$\pm$ 4.33}& 6.03 & \underline{0.71} $M$ \\

\rowcolor{ABC}
& 64 & \underline{0.75} \tiny{$\pm$ 0.17}  & \underline{1.96 }\tiny{$\pm$ 0.88}& \underline{1.21} & 0.75 $M$ \\

\rowcolor{ABC}
& 128 & 0.89 \tiny{$\pm$ 0.32}  & 2.48 \tiny{$\pm$ 0.67}& 1.59 & 0.84 $M$ \\

\rowcolor{ABC}
& 256 & \underline{0.75} \tiny{$\pm$ 0.23}  & 3.09 \tiny{$\pm$ 1.83}& 2.34 & 1.02 $M$ \\

 \hline

\hline
 
\hline
\end{tabular}}
\end{center}
\vskip -0.1in
\end{table}

\endgroup

\section{Additional Comparisons}

In this section, we expand our comparisons to include a broader range of state-of-the-art methods, as detailed below.

\paragraph{sys-bAbI task.}
We compare \texttt{D3} to state-of-the-art methods (DAM \cite{park2021distributed} and STM \cite{le2020self}) on the original bAbI task. Table~\ref{table.babi.sota} shows that existing memory networks struggle with the sys-bAbI task, highlighting the efficacy of \texttt{D3} compared to these state-of-the-art memory networks.

\begingroup
\setlength{\tabcolsep}{4pt} % Default value: 6pt
\renewcommand{\arraystretch}{1.2} % Default value: 1

\begin{table}[h]
\caption{The mean word error rate [\%] on additional comparison of the sys-bAbI task for 10 seeds.}
%\vskip -0.1in
\label{table.babi.sota}
\begin{center}
%{\scriptsize%\footnotesize
{\footnotesize
\begin{tabular}{l?l|l?c}
\hline

\hline

\hline
\bf Model            & \textit{w/o sys diff} ($\downarrow$)   &  \textit{w/ sys diff} ($\downarrow$)     & \textbf{Gap} ($\downarrow$)  \\ \hline

\hline

DAM     & 0.48 \tiny{$\pm$ 0.20}   & 5.25 \tiny{$\pm$ 1.64}    & 4.77     \\
STM     & 0.49 \tiny{$\pm$ 0.16}   & 4.79 \tiny{$\pm$ 1.53}    & 3.70     \\

\hline

\hline

TPR-RNN & 0.79 \tiny{$\pm$ 0.16}   & 8.74 \tiny{$\pm$ 3.74}    & 7.95     \\

~~~~~~{+ \scriptsize{AID}} & \underline{0.69} \tiny{$\pm$ 0.08} & \underline{5.61} \tiny{$\pm$ 1.78}  & \underline{4.92}   \\
\rowcolor{ABC}
~~~~~~{+ \texttt{\textbf{D3}}} & \textbf{0.65} \tiny{$\pm$ 0.25}  & \textbf{3.50} \tiny{$\pm$ 2.07}  & \textbf{2.85} \\

\hline

\hline

FWM     & 0.79 \tiny{$\pm$ 0.14}   & 2.85 \tiny{$\pm$ 1.61}    & 2.06     \\

 ~~~~~~{+ \scriptsize{AID}} & \textbf{0.45} \tiny{$\pm$ 0.16}  & \textbf{1.21} \tiny{$\pm$ 0.66}  & \textbf{0.76}   \\
 
\rowcolor{ABC}
 ~~~~~~{+ \texttt{\textbf{D3}} (\textit{w/o F})} & 0.79 \tiny{$\pm$ 0.30}  & 2.58 \tiny{$\pm$ 1.12}  & 1.79   \\

\rowcolor{ABC}
 ~~~~~~{+ \texttt{\textbf{D3}} (\textit{w/ F})} & \underline{0.75} \tiny{$\pm$ 0.17}  & \underline{1.96 }\tiny{$\pm$ 0.88}& \underline{1.21} \\

 \hline

\hline
 
\hline
\end{tabular}}
\end{center}
\end{table}

\endgroup

\paragraph{Sort-of-CLEVR task.}
We compare \texttt{D3} to vanilla Transformer \cite{vaswani2017attention} and Compositional Transformer \cite{mittal2021compositional}, designed to enhance the systematic generalization capabilities of multi-head self-attention methods. Table~\ref{table.clevr.sota} shows that the Linear Transformer significantly degrades systematic generalization performance compared to the vanilla Transformer and the Compositional Transformer. While \texttt{D3} improves the performance of the Linear Transformer from a TPR perspective, it still shows limited performance in reasoning the relationships between multiple objects (\textit{Binary} and \textit{Ternary}) compared to the vanilla Transformer and Compositional Transformer.

\begingroup
\setlength{\tabcolsep}{4pt} % Default value: 6pt
\renewcommand{\arraystretch}{1.2} % Default value: 1

\begin{table}[h]
%\vskip -0.1in
\caption{The mean accuracy [\%] on additional comparison of the sort-of-CLEVR task for 10 seeds.}
%\vskip -0.1in
\label{table.clevr.sota}
\begin{center}
{\footnotesize
\begin{tabular}{l?c?l|l|l}
\hline

\hline

\hline
 \bf Model       & $D_\text{code}$ & \textit{Unary} ($\uparrow$)~ & \textit{Binary} ($\uparrow$)  &  \textit{Ternary} ($\uparrow$)    \\ \hline 
 
 \hline

Transformer  &  - & 97.4 \tiny{$\pm$ 3.5}   & \underline{84.3} \tiny{$\pm$ 4.3}   &  62.7 \tiny{$\pm$ 3.9}   \\
Compositional Transformer  &  - & \underline{98.9} \tiny{$\pm$ 0.2}   & \textbf{88.4} \tiny{$\pm$ 1.4}   &  \underline{66.5} \tiny{$\pm$ 1.9}   \\

\hline

\hline

Linear Transformer  &  - & 69.3 \tiny{$\pm$ 14.8}   & 75.5 \tiny{$\pm$ 1.3}   &  56.4 \tiny{$\pm$ 4.3}   \\

~~~~~~~~~~{+ \scriptsize{AID}} &  - & \underline{98.9} \tiny{$\pm$ 0.2} & 78.6 \tiny{$\pm$ 0.3}  & 63.7 \tiny{$\pm$ 1.2}   \\

\hline

\rowcolor{ABC}
~~~~~~~~~~{+ \texttt{\textbf{D3}} (\textit{w/o F})} & 128 & 73.9 \tiny{$\pm$ 16.5} & 77.2 \tiny{$\pm$ 2.2}  & 57.3 \tiny{$\pm$ 4.6}   \\

\rowcolor{ABC}
 & 256 & 73.7 \tiny{$\pm$ 16.5}  & 77.8 \tiny{$\pm$ 2.5} & 57.9 \tiny{$\pm$ 5.8}   \\

\hline

\rowcolor{ABC}
~~~~~~~~~~{+ \texttt{\textbf{D3}} (\textit{w/ F})} & 128 & \underline{98.9} \tiny{$\pm$ 0.2}  & 79.5 \tiny{$\pm$ 0.8}  & 63.1 \tiny{$\pm$ 1.9}   \\

\rowcolor{ABC}
 & 256 & \textbf{99.0} \tiny{$\pm$ 0.3}  & 82.1 \tiny{$\pm$ 2.4}  & \textbf{68.8} \tiny{$\pm$ 1.2}   \\

 \hline
 
\hline

\hline
\end{tabular}}
\end{center}
\vskip -0.2in
\end{table}

\endgroup

\paragraph{WikiText-103 task.}
We compared \texttt{D3} to the Delta Network \cite{schlag2021linear}, which introduced a delta updating rule instead of the additive outer product-based updating rule in the Linear Transformer. Table~\ref{table.wiki.sota} indicates that although \texttt{D3} improves the performance of the Linear Transformer in language modeling tasks, the choice of updating rules has a more substantial impact on performance for tasks involving the comprehension of lengthy corpora than the decomposition operation.

\begingroup
\setlength{\tabcolsep}{4pt} % Default value: 6pt
\renewcommand{\arraystretch}{1.2} % Default value: 1

\begin{table}[h]
\vskip -0.1in
\caption{Perplexity on additional comparison of the WikiText-103 task.}
\label{table.wiki.sota}
\begin{center}
{\footnotesize
\begin{tabular}{l?c?c|c}
\hline

\hline

\hline
 \bf{Model}      & $D_\text{code}$   & \textit{Valid} ($\downarrow$)~  &  \textit{Test} ($\downarrow$)~~  \\ \hline
 
 \hline

\text{Delta Network}    & - & \textbf{35.640}   & \textbf{36.659}       \\

\hline

\hline

\text{Linear Transformer}    & - & 36.473   & 37.533       \\

~~~~~~~~~~ {+ \scriptsize{AID}} & - & 36.159  & {37.151}  \\

\hline

\rowcolor{ABC}
~~~~~~~~~~ {+ \texttt{\textbf{D3}} (\textit{w/o F})} & 32 & 36.061  & 37.220  \\

\rowcolor{ABC}
 & 64 & \underline{35.975} & \underline{37.009} \\

\hline

\rowcolor{ABC}
~~~~~~~~~~ {+ \texttt{\textbf{D3}} (\textit{w/ F})} & 32 & 36.630  & 37.620  \\

\rowcolor{ABC}
 & 64 & 36.220  & 37.128\\

 \hline
 
\hline

\hline
\end{tabular}}
\end{center}
\vskip -0.1in
\end{table}

\endgroup

\section{Additional Ablation Study}

In this section, we extend our ablation studies to investigate the effects of varying the number of keys in the codebook and the impact of removing either the residual connection or the codebook from the \texttt{D3} layer.

\paragraph{The Effect of Varying the Number of Codebook Keys.}

\begin{figure}[h]
\vskip -0.05in
\begin{center}
\centering
\includegraphics[width=0.7\linewidth]{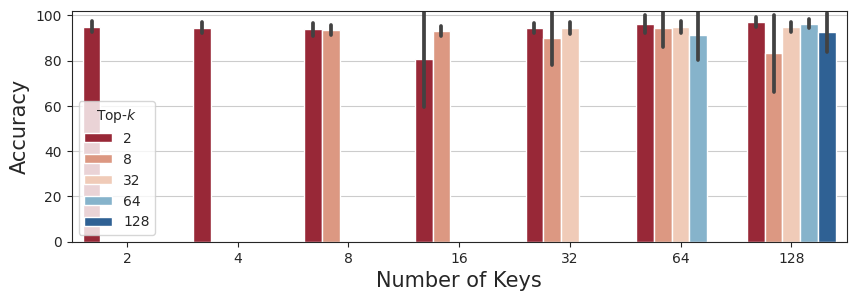}
\caption{
The mean accuracy on the SAR task for 10 seeds in the ablation study for the effect of varying $N_\text{code}$ from 2 to 128 with top-$k$ constant.
\label{fig.ablation.varying_key}}
\end{center}
\vskip -0.1in
\end{figure}

Fig.~\ref{fig.ablation.varying_key} shows that even with a significantly reduced number of keys, the model with \texttt{D3} maintains high accuracy on the SAR task. This observation prompts the question of how consistent performance is achieved despite the reduction in codebook size. To explore this further, we examine the impact of removing the codebook or the residual connection within the \texttt{D3} layer on the SAR and sys-bAbI tasks. Specifically, removing the codebook means that the \texttt{components} are generated solely by the shared feed-forward networks ($\text{layer}_\text{residual}$ and $\text{layer}_\text{final}$) while removing the residual connection implies that the \texttt{components} are derived solely from the codebook values.

\paragraph{The Effect of Residual Connection.}

\begin{figure}[h]
\vskip -0.05in
\begin{center}
\begin{subfigure}{.4\textwidth}
  \centering
  \includegraphics[width=\linewidth]{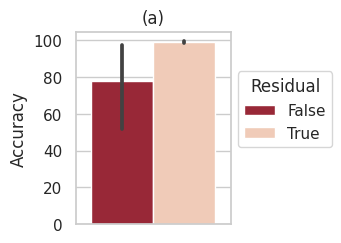}
\end{subfigure}
\begin{subfigure}{.59\textwidth}
  \centering
  \includegraphics[width=0.9\linewidth]{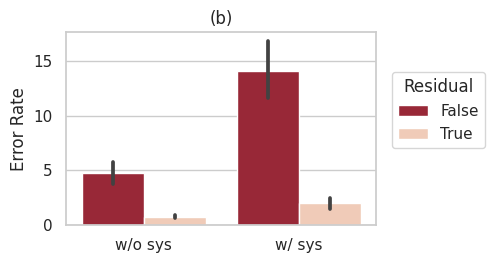}
\end{subfigure}
\caption{
Ablation study for the effect of the residual connection on (a) the SAR task and (b) the sys-bAbI task for 10 seeds.\label{fig.ablation.residual}
}
\end{center}
\vskip -0.1in
\end{figure}

Fig.~\ref{fig.ablation.residual} shows that without the residual connection, the generalization performance of \texttt{D3} dramatically degrades. This result indicates that the residual connection is crucial for effectively training the \texttt{D3} layer.

\paragraph{The Effect of Codebook.}
Table~\ref{table.sar.ablation.codebook} shows that even without the codebook ("\textit{w/o codebook}"), the \texttt{D3} layer improves the generalization performance of the baseline model on the SAR task. This result indicates that the shared feed-forward networks significantly contribute to performance enhancement, which may explain why the model maintains robust performance even with fewer keys.

However, it is important to note that without the codebook, the \texttt{D3} layer does not achieve near-perfect accuracy on the SAR task (as shown in Table~\ref{table.sar.ablation.codebook}) and fails to significantly enhance the systematic generalization of the baseline model on the sys-bAbI task (as shown in Table~\ref{table.babi.ablation.codebook}). These results demonstrate that the codebook plays a crucial role in enhancing the model's overall performance and generalization capabilities, especially in tasks requiring systematic generalization.

Furthermore, we experiment with $N_\text{code}=1$ on the SAR task, where the codebook may act as a bias term. The results in Table~\ref{table.sar.ablation.codebook}) show that using a single codebook element leads to degraded generalization performance compared to the "\textit{w/o codebook}" configuration, indicating that multiple codebook elements are essential for achieving optimal results.

\begin{table}[!htb]
\centering
\begin{minipage}[t]{\textwidth}
\centering
    \caption{Ablation study for the effect of the codebook on the SAR task for 10 seeds.\label{table.sar.ablation.codebook}}

\begingroup
\setlength{\tabcolsep}{4pt} % Default value: 6pt
\renewcommand{\arraystretch}{1.2} % Default value: 1

\begin{center}
%{\scriptsize%\footnotesize
{\footnotesize
\begin{tabular}{l?c|c|c?l}
\hline

\hline

\hline
\bf Model            & $D_\text{code}$   &  $N_\text{code}$   &  top-$k$  & \textbf{Accuracy} ($\uparrow$)  \\ \hline

\hline

FWM     & -  &  -  &  -  & ~~44.90 \tiny{$\pm$ 31.5}   \\

\hline
 
\rowcolor{ABC}
 ~~~~~~{+ \texttt{\textbf{D3}}} &   &  4  &  2  & ~~87.38 \tiny{$\pm$ 11.10}   \\

\rowcolor{ABC}
 ~~~~~~{+ \texttt{\textbf{D3}}} &   &  64  &  8  & ~~\textbf{99.27} \tiny{$\pm$ 0.88}   \\

\rowcolor{ABC}
 ~~~~~~{+ \texttt{\textbf{D3}} (\textit{w/o codebook})} & \multirow{-3}{*}{32}  &  -  &  -  & ~~\underline{89.02} \tiny{$\pm$ 4.56}   \\

\hline

\rowcolor{ABC}
 ~~~~~~{+ \texttt{\textbf{D3}}} &   &  1  &  1  & ~~89.10 \tiny{$\pm$ 7.99}   \\

\rowcolor{ABC}
 ~~~~~~{+ \texttt{\textbf{D3}}} &   &  4  &  2  & ~~\textbf{94.47} \tiny{$\pm$ 2.35}   \\

\rowcolor{ABC}
 ~~~~~~{+ \texttt{\textbf{D3}}} &   &  64  &  8  & ~~\underline{94.29} \tiny{$\pm$ 8.06}   \\

\rowcolor{ABC}
 ~~~~~~{+ \texttt{\textbf{D3}} (\textit{w/o codebook})} & \multirow{-4}{*}{64}  &  -  &  -  & ~~91.65 \tiny{$\pm$ 3.66}   \\

 \hline

\hline
 
\hline
\end{tabular}}
\end{center}

\endgroup

\end{minipage}%

\medskip{}

\begin{minipage}[t]{\textwidth}

\centering
    \caption{Ablation study for the effect of the codebook on the sys-bAbI task for 10 seeds.\label{table.babi.ablation.codebook}}

\begingroup
\setlength{\tabcolsep}{4pt} % Default value: 6pt
\renewcommand{\arraystretch}{1.2} % Default value: 1

\begin{center}
%{\scriptsize%\footnotesize
{\footnotesize
\begin{tabular}{l?l|l?c}
\hline

\hline

\hline
\bf Model            & \textit{w/o sys diff} ($\downarrow$)   &  \textit{w/ sys diff} ($\downarrow$)     & \textbf{Gap} ($\downarrow$)  \\ \hline

\hline

FWM     & \underline{0.79} \tiny{$\pm$ 0.14}   & \underline{2.85} \tiny{$\pm$ 1.61}    & \underline{2.06}     \\

\rowcolor{ABC}
 ~~~~~~{+ \texttt{\textbf{D3}}} & \textbf{0.75} \tiny{$\pm$ 0.17}  & \textbf{1.96 }\tiny{$\pm$ 0.88}& \textbf{1.21}   \\

\rowcolor{ABC}
 ~~~~~~{+ \texttt{\textbf{D3}} (\textit{w/o codebook})} & 1.19 \tiny{$\pm$ 0.41}  & 3.55\tiny{$\pm$ 1.04}& 2.36 \\

 \hline

\hline
 
\hline
\end{tabular}}
\end{center}

\endgroup

\end{minipage}

\end{table}

\paragraph{Discussion.}
Our ablation study on the codebook in the SAR task (Table~\ref{table.sar.ablation.codebook}) indicates that the shared residual networks within the \texttt{D3} layer significantly enhance generalization performance. However, the results from the sys-bAbI task (Table~\ref{table.babi.ablation.codebook}) suggest that while these networks improve performance, they alone struggle to generalize more structured data.

The ablation studies in Tables~\ref{table.sar.ablation.codebook} and~\ref{table.babi.ablation.codebook} demonstrate that incorporating the codebook mechanism leads to nearly 100\% accuracy on the SAR task and significantly improves the systematic generalization of models on the sys-bAbI task. However, as shown in the ablation study on the residual connection (Fig.~\ref{fig.ablation.residual}), the codebook alone does not achieve the same level of generalization and exhibits instability within the \texttt{D3} layer.

In conclusion, our experimental results indicate that the combination of the codebook and the shared residual networks within the \texttt{D3} layer is crucial for enhancing systematic generalization performance and stability. By integrating these two components, our \texttt{D3} layer significantly improves the systematic generalization capabilities of TPR-based models.

\section{Additional Qualitative Analysis}

\subsection{Comparison to Baselines}
\begin{figure}[h]
\begin{center}
\begin{minipage}[t]{0.9\linewidth}
\includegraphics[width=\linewidth]{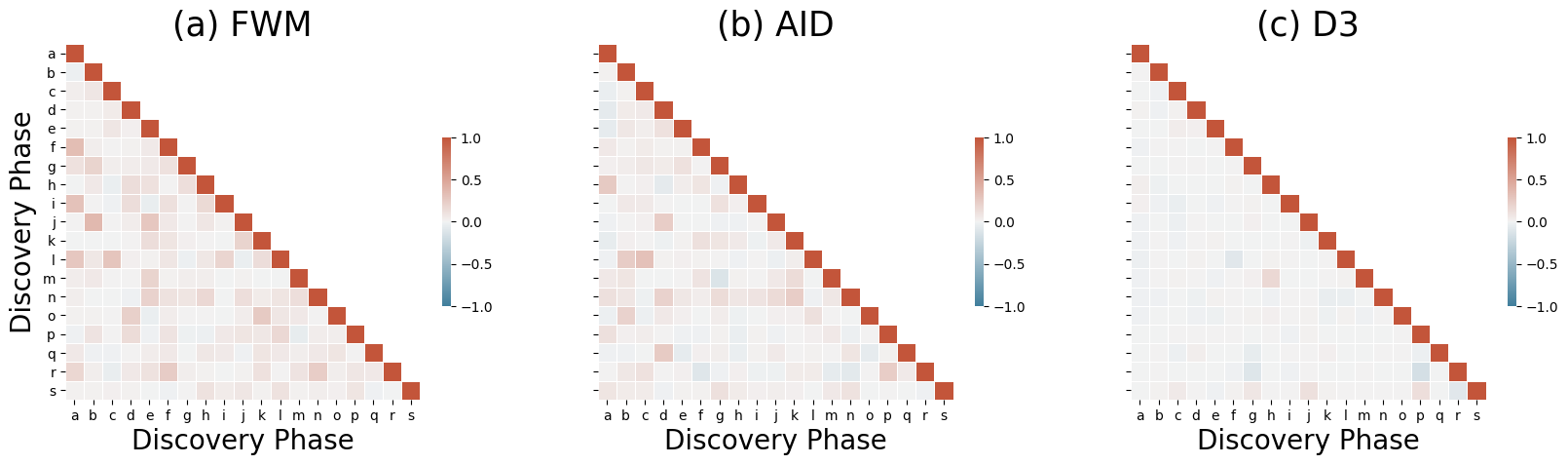}
\caption{
The heatmap displays the cosine similarity between the \textit{roles} during the discovery phase for the SAR task.\label{fig.sar.baseline.analysis.role}
}
\end{minipage}
\vskip 0.2in
\begin{minipage}[t]{\linewidth}
\centering
\includegraphics[width=0.9\linewidth]{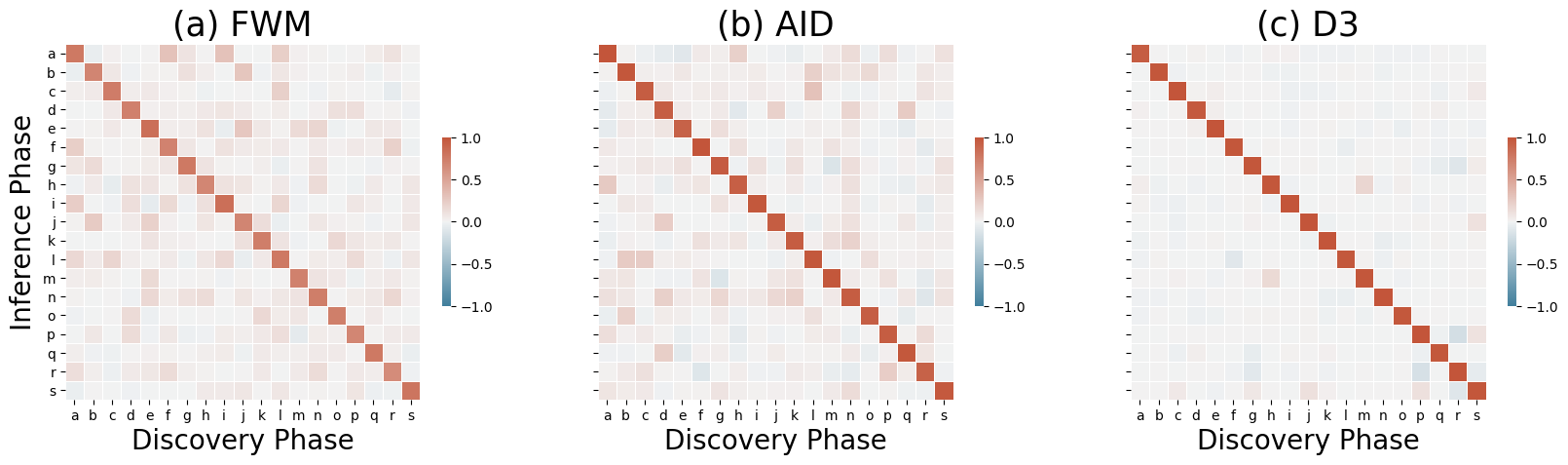}
\caption{
The heatmap displays the cosine similarity between the \textit{roles} (\textbf{x-axis}) during the discovery phase and the \textit{unbinding operators} (\textbf{y-axis}) during the inference phase for the SAR task.\label{fig.sar.baseline.analysis.role.unbind}
}
\end{minipage}
\end{center}
\end{figure}

We conduct an orthogonal analysis for the baseline models (FWM and AID) similar to the analysis presented in Section~\ref{sec.orthogonality}. Figs.~\ref{fig.sar.baseline.analysis.role} and~\ref{fig.sar.baseline.analysis.role.unbind} indicate that the D3 model generates more structured and orthogonal representations than the baseline models, FWM and AID, demonstrating its effectiveness.

\newpage

\subsection{Qualitative Analysis for Different Seeds}
Additionally, we present the results of the qualitative analysis for different seeds in the SAR task.

\subsubsection{$N_\text{code}$: 64, $D_\text{code}$: 32, top-$k$: 8, seed: 3333}

\begin{figure}[h]
\begin{center}
\begin{minipage}[t]{0.9\linewidth}
\includegraphics[width=\linewidth]{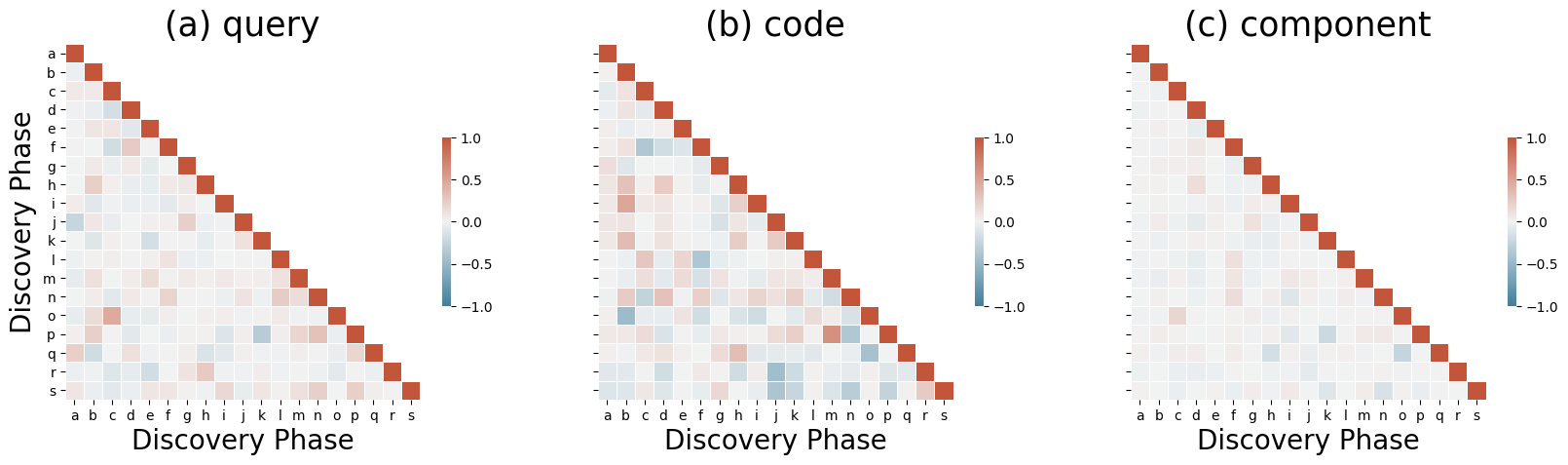}
\caption{
The heatmap displays the cosine similarity between the generated representations during the discovery phase for the SAR task. We explore the similarity across different types of representations: (a) \texttt{queries} of \textit{roles}, (b) \texttt{codes} of \textit{roles}, and (c) the \textit{roles} themselves.
}
\end{minipage}
\vskip 0.2in
\begin{minipage}[t]{0.9\linewidth}
\includegraphics[width=\linewidth]{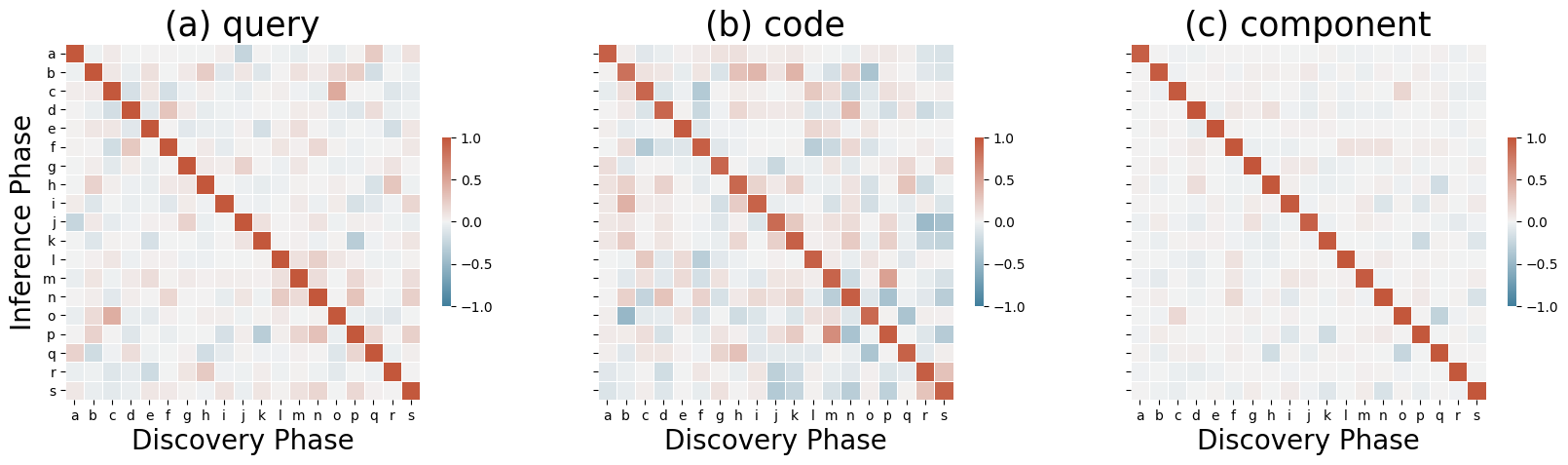}
\caption{
The heatmap displays the cosine similarity between the generated representations during the discovery phase (represented on the \textbf{x-axis}) and the inference phase (represented on the \textbf{y-axis}) for the SAR task. We explore the similarity across different types of representations: (a) \texttt{queries} of \textit{roles} and \textit{unbinding operators}, (b) \texttt{codes} of \textit{roles} and \textit{unbinding operators}, and (c) the \textit{roles} and \textit{unbinding operators} themselves.
}
\end{minipage}
\vskip 0.2in
\begin{minipage}[t]{0.9\linewidth}
\includegraphics[width=\linewidth]{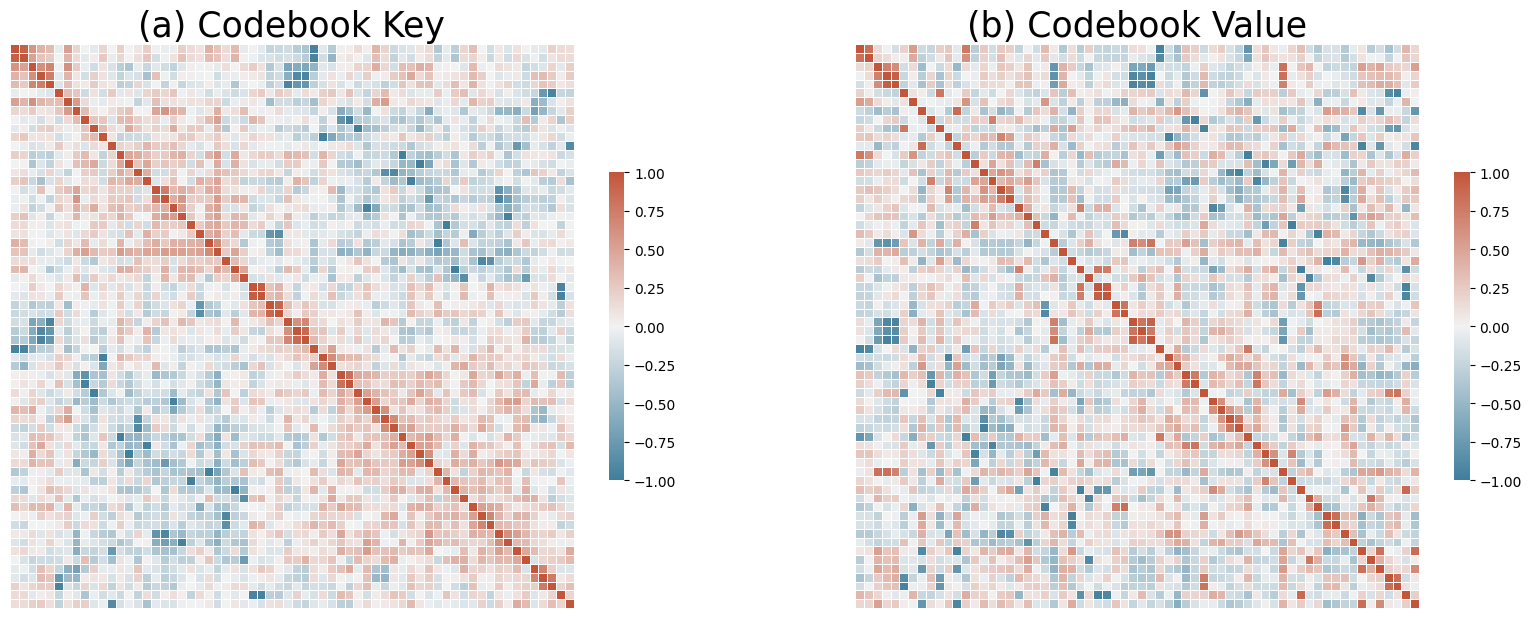}
\caption{
The heatmap visualizes the cosine similarity of the learned codebook features for the SAR task. There are two parts to each heatmap: (a) the similarity among codebook keys, denoted as $\{\textsf{k}_i\}_{i=1}^{N\text{code}}$, and (b) the similarity among codebook values, denoted as $\{\textsf{v}_i\}_{i=1}^{N\text{code}}$. For better visualization, the heatmap values are reordered to reflect the cluster of similar codebook keys.
}
\end{minipage}
\end{center}
\end{figure}

\newpage

\subsubsection{$N_\text{code}$: 64, $D_\text{code}$: 32, top-$k$: 8, seed: 4444}

\begin{figure}[h]
\begin{center}
\begin{minipage}[t]{0.9\linewidth}
\includegraphics[width=\linewidth]{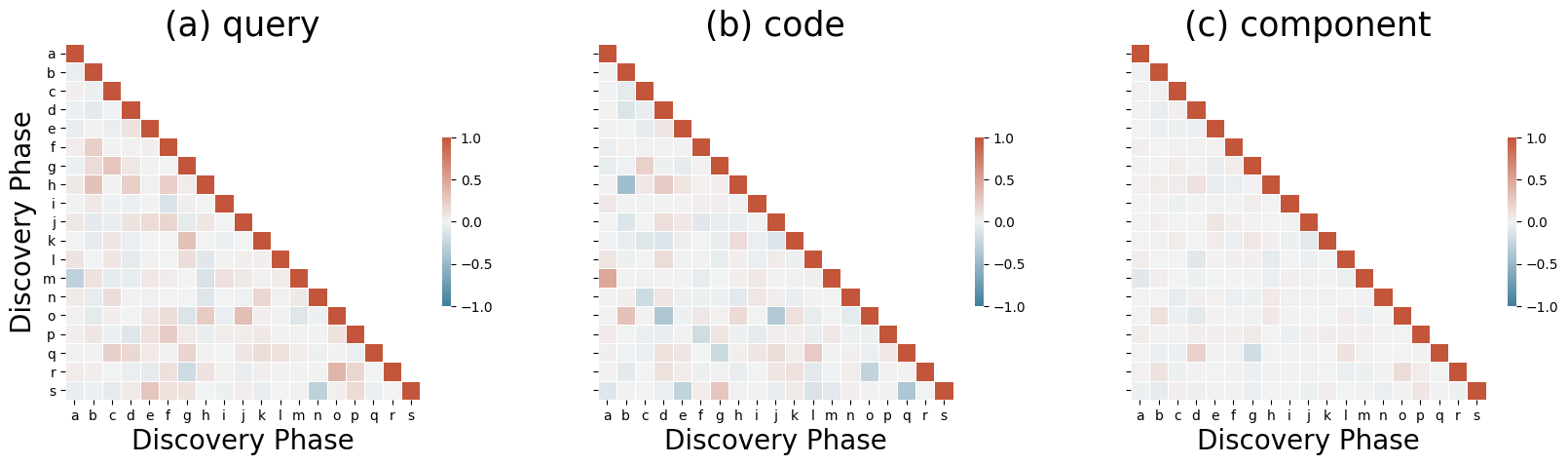}
\caption{
The heatmap displays the cosine similarity between the generated representations during the discovery phase for the SAR task. We explore the similarity across different types of representations: (a) \texttt{queries} of \textit{roles}, (b) \texttt{codes} of \textit{roles}, and (c) the \textit{roles} themselves.
}
\end{minipage}
\vskip 0.2in
\begin{minipage}[t]{0.9\linewidth}
\includegraphics[width=\linewidth]{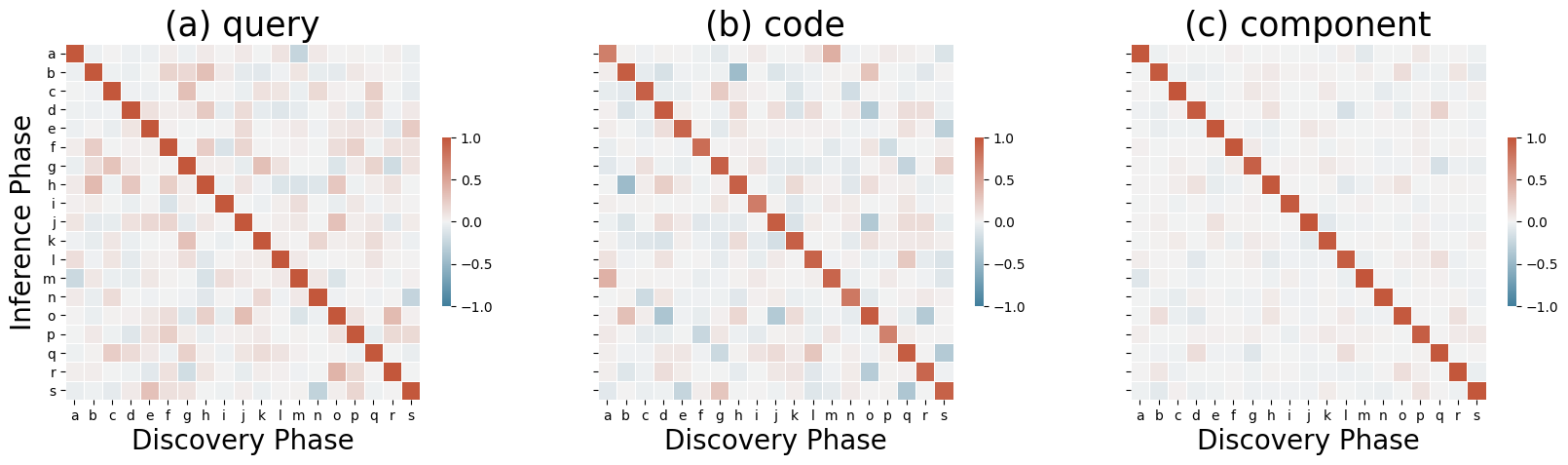}
\caption{
The heatmap displays the cosine similarity between the generated representations during the discovery phase (represented on the \textbf{x-axis}) and the inference phase (represented on the \textbf{y-axis}) for the SAR task. We explore the similarity across different types of representations: (a) \texttt{queries} of \textit{roles} and \textit{unbinding operators}, (b) \texttt{codes} of \textit{roles} and \textit{unbinding operators}, and (c) the \textit{roles} and \textit{unbinding operators} themselves.
}
\end{minipage}
\vskip 0.2in
\begin{minipage}[t]{0.9\linewidth}
\includegraphics[width=\linewidth]{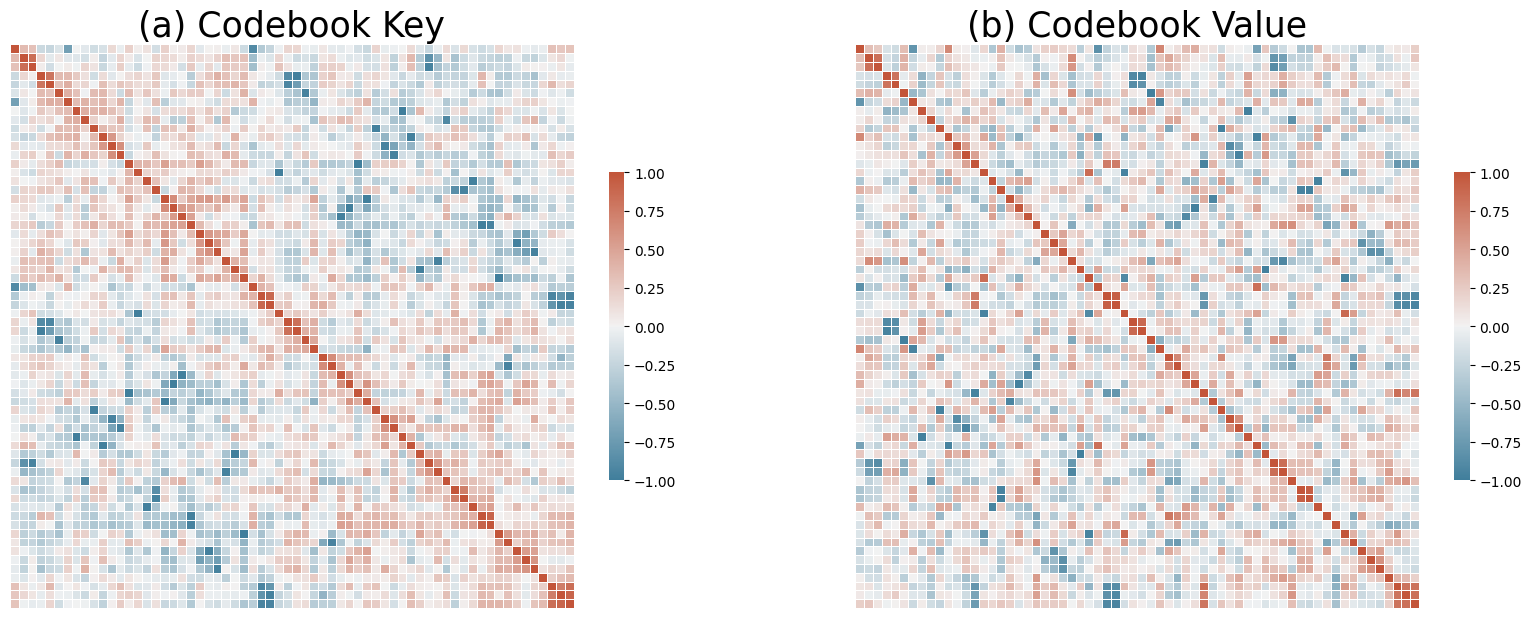}
\caption{
The heatmap visualizes the cosine similarity of the learned codebook features for the SAR task. There are two parts to each heatmap: (a) the similarity among codebook keys, denoted as $\{\textsf{k}_i\}_{i=1}^{N\text{code}}$, and (b) the similarity among codebook values, denoted as $\{\textsf{v}_i\}_{i=1}^{N\text{code}}$. For better visualization, the heatmap values are reordered to reflect the cluster of similar codebook keys.
}
\end{minipage}
\end{center}
\end{figure}

\end{appendices}

\pagebreak

\section*{NeurIPS Paper Checklist}

\begin{enumerate}

\item {\bf Claims}
    \item[] Question: Do the main claims made in the abstract and introduction accurately reflect the paper's contributions and scope?
    \item[] Answer: \answerYes{} % Replace by \answerYes{}, \answerNo{}, or \answerNA{}.
    \item[] Justification: This paper includes the paper's contributions and scope in the abstract and introduction, as follows. This paper tackles the decomposition problem inherent in the TPR-based approaches. To address this, this paper proposes a discrete dictionary-based decomposition (D3) layer designed to enhance the decomposition capabilities of the TPR-based models.
    \item[] Guidelines:
    \begin{itemize}
        \item The answer NA means that the abstract and introduction do not include the claims made in the paper.
        \item The abstract and/or introduction should clearly state the claims made, including the contributions made in the paper and important assumptions and limitations. A No or NA answer to this question will not be perceived well by the reviewers. 
        \item The claims made should match theoretical and experimental results, and reflect how much the results can be expected to generalize to other settings. 
        \item It is fine to include aspirational goals as motivation as long as it is clear that these goals are not attained by the paper. 
    \end{itemize}

\item {\bf Limitations}
    \item[] Question: Does the paper discuss the limitations of the work performed by the authors?
    \item[] Answer: \answerYes{} % Replace by \answerYes{}, \answerNo{}, or \answerNA{}.
    \item[] Justification: This paper discusses the limitations of the work performed by the authors in Section~\ref{sec.discussion}, as follows. The model introduced in this paper requires additional computational overhead and configuration search when the proposed model is integrated into the existing baseline models.
    \item[] Guidelines:
    \begin{itemize}
        \item The answer NA means that the paper has no limitation while the answer No means that the paper has limitations, but those are not discussed in the paper. 
        \item The authors are encouraged to create a separate "Limitations" section in their paper.
        \item The paper should point out any strong assumptions and how robust the results are to violations of these assumptions (e.g., independence assumptions, noiseless settings, model well-specification, asymptotic approximations only holding locally). The authors should reflect on how these assumptions might be violated in practice and what the implications would be.
        \item The authors should reflect on the scope of the claims made, e.g., if the approach was only tested on a few datasets or with a few runs. In general, empirical results often depend on implicit assumptions, which should be articulated.
        \item The authors should reflect on the factors that influence the performance of the approach. For example, a facial recognition algorithm may perform poorly when image resolution is low or images are taken in low lighting. Or a speech-to-text system might not be used reliably to provide closed captions for online lectures because it fails to handle technical jargon.
        \item The authors should discuss the computational efficiency of the proposed algorithms and how they scale with dataset size.
        \item If applicable, the authors should discuss possible limitations of their approach to address problems of privacy and fairness.
        \item While the authors might fear that complete honesty about limitations might be used by reviewers as grounds for rejection, a worse outcome might be that reviewers discover limitations that aren't acknowledged in the paper. The authors should use their best judgment and recognize that individual actions in favor of transparency play an important role in developing norms that preserve the integrity of the community. Reviewers will be specifically instructed to not penalize honesty concerning limitations.
    \end{itemize}

\item {\bf Theory Assumptions and Proofs}
    \item[] Question: For each theoretical result, does the paper provide the full set of assumptions and a complete (and correct) proof?
    \item[] Answer: \answerNA{} % Replace by \answerYes{}, \answerNo{}, or \answerNA{}.
    \item[] Justification: This paper does not include theoretical results.
    \item[] Guidelines:
    \begin{itemize}
        \item The answer NA means that the paper does not include theoretical results. 
        \item All the theorems, formulas, and proofs in the paper should be numbered and cross-referenced.
        \item All assumptions should be clearly stated or referenced in the statement of any theorems.
        \item The proofs can either appear in the main paper or the supplemental material, but if they appear in the supplemental material, the authors are encouraged to provide a short proof sketch to provide intuition. 
        \item Inversely, any informal proof provided in the core of the paper should be complemented by formal proofs provided in appendix or supplemental material.
        \item Theorems and Lemmas that the proof relies upon should be properly referenced. 
    \end{itemize}

    \item {\bf Experimental Result Reproducibility}
    \item[] Question: Does the paper fully disclose all the information needed to reproduce the main experimental results of the paper to the extent that it affects the main claims and/or conclusions of the paper (regardless of whether the code and data are provided or not)?
    \item[] Answer: \answerYes{} % Replace by \answerYes{}, \answerNo{}, or \answerNA{}.
    \item[] Justification: This paper discloses all the information needed to reproduce the experimental results. This paper explains the mechanism of the proposed model and how it is applied to existing baseline models in Section~\ref{sec.method} and presents the experiment details and hyper-parameter settings in Appendices~\ref{appendix.experiment.design} and~\ref{appendix.experiment.parameter}.
    \item[] Guidelines:
    \begin{itemize}
        \item The answer NA means that the paper does not include experiments.
        \item If the paper includes experiments, a No answer to this question will not be perceived well by the reviewers: Making the paper reproducible is important, regardless of whether the code and data are provided or not.
        \item If the contribution is a dataset and/or model, the authors should describe the steps taken to make their results reproducible or verifiable. 
        \item Depending on the contribution, reproducibility can be accomplished in various ways. For example, if the contribution is a novel architecture, describing the architecture fully might suffice, or if the contribution is a specific model and empirical evaluation, it may be necessary to either make it possible for others to replicate the model with the same dataset, or provide access to the model. In general. releasing code and data is often one good way to accomplish this, but reproducibility can also be provided via detailed instructions for how to replicate the results, access to a hosted model (e.g., in the case of a large language model), releasing of a model checkpoint, or other means that are appropriate to the research performed.
        \item While NeurIPS does not require releasing code, the conference does require all submissions to provide some reasonable avenue for reproducibility, which may depend on the nature of the contribution. For example
        \begin{enumerate}
            \item If the contribution is primarily a new algorithm, the paper should make it clear how to reproduce that algorithm.
            \item If the contribution is primarily a new model architecture, the paper should describe the architecture clearly and fully.
            \item If the contribution is a new model (e.g., a large language model), then there should either be a way to access this model for reproducing the results or a way to reproduce the model (e.g., with an open-source dataset or instructions for how to construct the dataset).
            \item We recognize that reproducibility may be tricky in some cases, in which case authors are welcome to describe the particular way they provide for reproducibility. In the case of closed-source models, it may be that access to the model is limited in some way (e.g., to registered users), but it should be possible for other researchers to have some path to reproducing or verifying the results.
        \end{enumerate}
    \end{itemize}

\item {\bf Open access to data and code}
    \item[] Question: Does the paper provide open access to the data and code, with sufficient instructions to faithfully reproduce the main experimental results, as described in supplemental material?
    \item[] Answer: \answerYes{} % Replace by \answerYes{}, \answerNo{}, or \answerNA{}.
    \item[] Justification: This paper provides supplementary materials to reproduce all experimental results of the proposed method, including source codes about our model implementation, data processing, scripts for execution, etc.
    \item[] Guidelines:
    \begin{itemize}
        \item The answer NA means that paper does not include experiments requiring code.
        \item Please see the NeurIPS code and data submission guidelines (\url{https://nips.cc/public/guides/CodeSubmissionPolicy}) for more details.
        \item While we encourage the release of code and data, we understand that this might not be possible, so “No” is an acceptable answer. Papers cannot be rejected simply for not including code, unless this is central to the contribution (e.g., for a new open-source benchmark).
        \item The instructions should contain the exact command and environment needed to run to reproduce the results. See the NeurIPS code and data submission guidelines (\url{https://nips.cc/public/guides/CodeSubmissionPolicy}) for more details.
        \item The authors should provide instructions on data access and preparation, including how to access the raw data, preprocessed data, intermediate data, and generated data, etc.
        \item The authors should provide scripts to reproduce all experimental results for the new proposed method and baselines. If only a subset of experiments are reproducible, they should state which ones are omitted from the script and why.
        \item At submission time, to preserve anonymity, the authors should release anonymized versions (if applicable).
        \item Providing as much information as possible in supplemental material (appended to the paper) is recommended, but including URLs to data and code is permitted.
    \end{itemize}

\item {\bf Experimental Setting/Details}
    \item[] Question: Does the paper specify all the training and test details (e.g., data splits, hyperparameters, how they were chosen, type of optimizer, etc.) necessary to understand the results?
    \item[] Answer: \answerYes{} % Replace by \answerYes{}, \answerNo{}, or \answerNA{}.
    \item[] Justification: This paper presents our experiment details and hyper-parameter settings in Appendices~\ref{appendix.experiment.design} and~\ref{appendix.experiment.parameter}.
    \item[] Guidelines:
    \begin{itemize}
        \item The answer NA means that the paper does not include experiments.
        \item The experimental setting should be presented in the core of the paper to a level of detail that is necessary to appreciate the results and make sense of them.
        \item The full details can be provided either with the code, in appendix, or as supplemental material.
    \end{itemize}

\item {\bf Experiment Statistical Significance}
    \item[] Question: Does the paper report error bars suitably and correctly defined or other appropriate information about the statistical significance of the experiments?
    \item[] Answer: \answerYes{} % Replace by \answerYes{}, \answerNo{}, or \answerNA{}.
    \item[] Justification: This paper reports the mean and standard deviation values in the experimental results conducted using fixed 10 different random seeds.
    \item[] Guidelines:
    \begin{itemize}
        \item The answer NA means that the paper does not include experiments.
        \item The authors should answer "Yes" if the results are accompanied by error bars, confidence intervals, or statistical significance tests, at least for the experiments that support the main claims of the paper.
        \item The factors of variability that the error bars are capturing should be clearly stated (for example, train/test split, initialization, random drawing of some parameter, or overall run with given experimental conditions).
        \item The method for calculating the error bars should be explained (closed form formula, call to a library function, bootstrap, etc.)
        \item The assumptions made should be given (e.g., Normally distributed errors).
        \item It should be clear whether the error bar is the standard deviation or the standard error of the mean.
        \item It is OK to report 1-sigma error bars, but one should state it. The authors should preferably report a 2-sigma error bar than state that they have a 96\% CI, if the hypothesis of Normality of errors is not verified.
        \item For asymmetric distributions, the authors should be careful not to show in tables or figures symmetric error bars that would yield results that are out of range (e.g. negative error rates).
        \item If error bars are reported in tables or plots, The authors should explain in the text how they were calculated and reference the corresponding figures or tables in the text.
    \end{itemize}

\item {\bf Experiments Compute Resources}
    \item[] Question: For each experiment, does the paper provide sufficient information on the computer resources (type of compute workers, memory, time of execution) needed to reproduce the experiments?
    \item[] Answer: \answerYes{} % Replace by \answerYes{}, \answerNo{}, or \answerNA{}.
    \item[] Justification: This paper provides the computer resources used in our experiments and the time it took to learn each task in Appendix~\ref{appendix.experiment.design}.
    \item[] Guidelines:
    \begin{itemize}
        \item The answer NA means that the paper does not include experiments.
        \item The paper should indicate the type of compute workers CPU or GPU, internal cluster, or cloud provider, including relevant memory and storage.
        \item The paper should provide the amount of compute required for each of the individual experimental runs as well as estimate the total compute. 
        \item The paper should disclose whether the full research project required more compute than the experiments reported in the paper (e.g., preliminary or failed experiments that didn't make it into the paper). 
    \end{itemize}
    
\item {\bf Code Of Ethics}
    \item[] Question: Does the research conducted in the paper conform, in every respect, with the NeurIPS Code of Ethics \url{https://neurips.cc/public/EthicsGuidelines}?
    \item[] Answer: \answerYes{} % Replace by \answerYes{}, \answerNo{}, or \answerNA{}.
    \item[] Justification: The research conducted in this paper conforms to the NeurIPS Code of Ethics.
    \item[] Guidelines:
    \begin{itemize}
        \item The answer NA means that the authors have not reviewed the NeurIPS Code of Ethics.
        \item If the authors answer No, they should explain the special circumstances that require a deviation from the Code of Ethics.
        \item The authors should make sure to preserve anonymity (e.g., if there is a special consideration due to laws or regulations in their jurisdiction).
    \end{itemize}

\item {\bf Broader Impacts}
    \item[] Question: Does the paper discuss both potential positive societal impacts and negative societal impacts of the work performed?
    \item[] Answer: \answerNA{} % Replace by \answerYes{}, \answerNo{}, or \answerNA{}.
    \item[] Justification: The authors do not foresee a negative societal impact on the work presented in this paper beyond the general effects of ML advancements.
    \item[] Guidelines:
    \begin{itemize}
        \item The answer NA means that there is no societal impact of the work performed.
        \item If the authors answer NA or No, they should explain why their work has no societal impact or why the paper does not address societal impact.
        \item Examples of negative societal impacts include potential malicious or unintended uses (e.g., disinformation, generating fake profiles, surveillance), fairness considerations (e.g., deployment of technologies that could make decisions that unfairly impact specific groups), privacy considerations, and security considerations.
        \item The conference expects that many papers will be foundational research and not tied to particular applications, let alone deployments. However, if there is a direct path to any negative applications, the authors should point it out. For example, it is legitimate to point out that an improvement in the quality of generative models could be used to generate deepfakes for disinformation. On the other hand, it is not needed to point out that a generic algorithm for optimizing neural networks could enable people to train models that generate Deepfakes faster.
        \item The authors should consider possible harms that could arise when the technology is being used as intended and functioning correctly, harms that could arise when the technology is being used as intended but gives incorrect results, and harms following from (intentional or unintentional) misuse of the technology.
        \item If there are negative societal impacts, the authors could also discuss possible mitigation strategies (e.g., gated release of models, providing defenses in addition to attacks, mechanisms for monitoring misuse, mechanisms to monitor how a system learns from feedback over time, improving the efficiency and accessibility of ML).
    \end{itemize}
    
\item {\bf Safeguards}
    \item[] Question: Does the paper describe safeguards that have been put in place for responsible release of data or models that have a high risk for misuse (e.g., pretrained language models, image generators, or scraped datasets)?
    \item[] Answer: \answerNA{} % Replace by \answerYes{}, \answerNo{}, or \answerNA{}.
    \item[] Justification: This paper does not have a high risk for misuse.
    \item[] Guidelines:
    \begin{itemize}
        \item The answer NA means that the paper poses no such risks.
        \item Released models that have a high risk for misuse or dual-use should be released with necessary safeguards to allow for controlled use of the model, for example by requiring that users adhere to usage guidelines or restrictions to access the model or implementing safety filters. 
        \item Datasets that have been scraped from the Internet could pose safety risks. The authors should describe how they avoided releasing unsafe images.
        \item We recognize that providing effective safeguards is challenging, and many papers do not require this, but we encourage authors to take this into account and make a best faith effort.
    \end{itemize}

\item {\bf Licenses for existing assets}
    \item[] Question: Are the creators or original owners of assets (e.g., code, data, models), used in the paper, properly credited and are the license and terms of use explicitly mentioned and properly respected?
    \item[] Answer: \answerYes{} % Replace by \answerYes{}, \answerNo{}, or \answerNA{}.
    \item[] Justification: This paper cites the original paper that produced the code package or dataset, and includes URLs in Appendix~\ref{appendix.experiment.design}.
    \item[] Guidelines:
    \begin{itemize}
        \item The answer NA means that the paper does not use existing assets.
        \item The authors should cite the original paper that produced the code package or dataset.
        \item The authors should state which version of the asset is used and, if possible, include a URL.
        \item The name of the license (e.g., CC-BY 4.0) should be included for each asset.
        \item For scraped data from a particular source (e.g., website), the copyright and terms of service of that source should be provided.
        \item If assets are released, the license, copyright information, and terms of use in the package should be provided. For popular datasets, \url{paperswithcode.com/datasets} has curated licenses for some datasets. Their licensing guide can help determine the license of a dataset.
        \item For existing datasets that are re-packaged, both the original license and the license of the derived asset (if it has changed) should be provided.
        \item If this information is not available online, the authors are encouraged to reach out to the asset's creators.
    \end{itemize}

\item {\bf New Assets}
    \item[] Question: Are new assets introduced in the paper well documented and is the documentation provided alongside the assets?
    \item[] Answer: \answerYes{} % Replace by \answerYes{}, \answerNo{}, or \answerNA{}.
    \item[] Justification: This paper provides supplementary materials with source code, license, and README.md files. The README.md files cite the code packages utilized in this paper and provide all the instructions to reproduce the experimental results.
    \item[] Guidelines:
    \begin{itemize}
        \item The answer NA means that the paper does not release new assets.
        \item Researchers should communicate the details of the dataset/code/model as part of their submissions via structured templates. This includes details about training, license, limitations, etc. 
        \item The paper should discuss whether and how consent was obtained from people whose asset is used.
        \item At submission time, remember to anonymize your assets (if applicable). You can either create an anonymized URL or include an anonymized zip file.
    \end{itemize}

\item {\bf Crowdsourcing and Research with Human Subjects}
    \item[] Question: For crowdsourcing experiments and research with human subjects, does the paper include the full text of instructions given to participants and screenshots, if applicable, as well as details about compensation (if any)? 
    \item[] Answer: \answerNA{} % Replace by \answerYes{}, \answerNo{}, or \answerNA{}.
    \item[] Justification:  This paper does not involve crowdsourcing nor research with human subjects.
    \item[] Guidelines:
    \begin{itemize}
        \item The answer NA means that the paper does not involve crowdsourcing nor research with human subjects.
        \item Including this information in the supplemental material is fine, but if the main contribution of the paper involves human subjects, then as much detail as possible should be included in the main paper. 
        \item According to the NeurIPS Code of Ethics, workers involved in data collection, curation, or other labor should be paid at least the minimum wage in the country of the data collector. 
    \end{itemize}

\item {\bf Institutional Review Board (IRB) Approvals or Equivalent for Research with Human Subjects}
    \item[] Question: Does the paper describe potential risks incurred by study participants, whether such risks were disclosed to the subjects, and whether Institutional Review Board (IRB) approvals (or an equivalent approval/review based on the requirements of your country or institution) were obtained?
    \item[] Answer: \answerNA{} % Replace by \answerYes{}, \answerNo{}, or \answerNA{}.
    \item[] Justification:  This paper does not involve crowdsourcing nor research with human subjects.
    \item[] Guidelines:
    \begin{itemize}
        \item The answer NA means that the paper does not involve crowdsourcing nor research with human subjects.
        \item Depending on the country in which research is conducted, IRB approval (or equivalent) may be required for any human subjects research. If you obtained IRB approval, you should clearly state this in the paper. 
        \item We recognize that the procedures for this may vary significantly between institutions and locations, and we expect authors to adhere to the NeurIPS Code of Ethics and the guidelines for their institution. 
        \item For initial submissions, do not include any information that would break anonymity (if applicable), such as the institution conducting the review.
    \end{itemize}

\end{enumerate}
%\appendix

% \section{Appendix / supplemental material}

% Optionally include supplemental material (complete proofs, additional experiments and plots) in appendix.
% All such materials \textbf{SHOULD be included in the main submission.}

%%%%%%%%%%%%%%%%%%%%%%%%%%%%%%%%%%%%%%%%%%%%%%%%%%%%%%%%%%%%

\end{document}